\documentclass[lettersize,journal]{IEEEtran}
\usepackage{amsmath,amsfonts}
\usepackage{subcaption}
\usepackage{algorithmic}
\usepackage{algorithm}
\usepackage{array}
\usepackage{threeparttable}
\usepackage{textcomp}
\usepackage{stfloats}
\usepackage{xcolor}
\usepackage{url}
\usepackage{verbatim}
\usepackage{graphicx}
\usepackage{subcaption}

\usepackage{cite}
\usepackage{makecell}
\usepackage[utf8]{inputenc} % 支持 UTF-8 编码
\usepackage{booktabs}       % 用于美观的表格线
\usepackage{array}          % 提供对表格列格式的更多控制
\usepackage{multirow,tabularx}       % 允许单元格跨多行

\newtheorem{problem}{PROBLEM}

\newtheorem{definition}{DEFINITION}
\newcommand{\para}[1]{{\vspace{2pt} \bf \noindent #1 \hspace{0.5pt}}}
\begin{document}

% \title{X-MLM: A Cross-City Learning Framework with LLM-Enhanced Intention Modeling for Mobility Prediction in extreme scenarios}
\title{Predicting Human Mobility during Extreme Events via LLM-Enhanced Cross-City Learning}

\author{Yinzhou Tang,
        Huandong Wang,
         % ~\IEEEmembership{Member, IEEE},
        Xiaochen Fan, 
        and Yong Li
         % ~\IEEEmembership{Senior Member, IEEE}% <-this % stops a space
\IEEEcompsocitemizethanks{\IEEEcompsocthanksitem 
Y. Tang, H. Wang, X. Fan, and Y. Li are with the Department of
Electronic Engineering, Beijing National Research Center
for Information Science and Technology (BNRist), Tsinghua University, Beijing 100084, China.
%(E-mail: tangyinzhou0131@163.com, 
%wanghuandong@mail.tsinghua.edu.cn,
%fanxiaochen33@gmail.com,
%liyong07@tsinghua.edu.cn).
\protect\\
}
}

\newenvironment{packed_itemize}{
\begin{list}{\labelitemi}{\leftmargin=1.5em}
  \setlength{\itemsep}{1pt}
  \setlength{\parskip}{0pt}
  \setlength{\parsep}{0pt}
  \setlength{\headsep}{0pt}
  \setlength{\topskip}{0pt}
  \setlength{\topmargin}{0pt}
  \setlength{\topsep}{0pt}
  \setlength{\partopsep}{0pt}
}{\end{list}}

% The paper headers
%\markboth{Journal of \LaTeX\ Class Files,~Vol.~14, No.~8, August~2021}%{Shell \MakeLowercase{\textit{et al.}}: A Sample Article Using IEEEtran.cls for IEEE Journals}

% \IEEEpubid{0000--0000/00\$00.00~\copyright~2021 IEEE}
% Remember, if you use this you must call \IEEEpubidadjcol in the second
% column for its text to clear the IEEEpubid mark.

\maketitle

\begin{abstract}
The vulnerability of cities has increased with urbanization and climate change, making it more important to predict human mobility during extreme events (e.g., extreme weather) for downstream tasks including location-based early disaster warning and pre-allocating rescue resources, etc.
% This is of great significance for post-disaster resource allocation, individual-level mobility warning and guiding the designated emergency response of the impending disaster city. 
However, existing human mobility prediction models are mainly designed for normal scenarios, and fail to adapt to extreme scenarios due to the shift of human mobility patterns under extreme scenarios.
% However, because the human mobility pattern in extreme scenarios is different from that in normal scenarios, it is difficult for the existing normal human mobility prediction models to handle it in extreme scenarios. 
% In addition, this task still faces the problem that the target city has little mobility data available under a particular disaster situation beyond experience.
To address this issue, we introduce \textbf{X-MLM}, a cross-e\textbf{X}treme-event \textbf{M}obility \textbf{L}anguge \textbf{M}odel framework for extreme scenarios that can be integrated into existing deep mobility prediction methods by leveraging LLMs to model the mobility intention and transferring the common knowledge of how different extreme events affect mobility intentions between cities. This framework utilizes a RAG-Enhanced Intention Predictor to forecast the next intention, refines it with an LLM-based Intention Refiner, and then maps the intention to an exact location using an Intention-Modulated Location Predictor.
% To address these issues, we designed a method that combines knowledge across cities and disasters at the intention level and uses LLMs to predict the mobility of an individual.
Extensive experiments illustrate that X-MLM can achieve a 32.8\% improvement in terms of Acc@1 and a 35.0\% improvement in terms of the F1-score of predicting immobility compared to the baselines. 
%It can also compensate for the performance decrease in mobility prediction models in extreme scenarios.
The code is available at \url{https://github.com/tsinghua-fib-lab/XMLM}.
\end{abstract}


\section{Introduction}\label{sec:introduction}
\IEEEPARstart{W}{ith} the rapid urbanization~\cite{sun2020dramatic} and climate change~\cite{leichenko2011climate}, cities across the world are becoming increasingly vulnerable to extreme events (e.g., extreme weather, traffic congestion), exposing more human lives and properties at risk.
To tackle these challenges, a fundamental research problem is to predict human mobility during extreme scenarios, which can support a wide spectrum of downstream emergency response tasks including
location-based early disaster warning~\cite{rahman2012location,zhao2022estimating,zhang2024agent},
pre-allocating rescue resources~\cite{tang2024predicting}, 
and planning humanitarian relief~\cite{song2016prediction}, etc.

 \iffalse
As natural disasters are becoming more and more frequent these days, it is getting more important to predict human mobility in extreme scenarios~\cite{RitchieRosadoRoser2022}. Accurately predicting the next location of an individual in extreme scenarios will benefit downstream tasks such as path planning for post-disaster resource allocation and evacuation~\cite{song2016prediction}, individual-level evacuation warning~\cite{mahiddin2021review}, and guiding the designated emergency response of the impending disaster city~\cite{fan2016citycoupling}.
\fi

\begin{figure}[t]
  \centering
  % 第一行
  \begin{subfigure}[b]{0.23\textwidth}  % [b] 表示基线对齐, {0.45\textwidth} 是子图宽度
    \includegraphics[width=\textwidth]{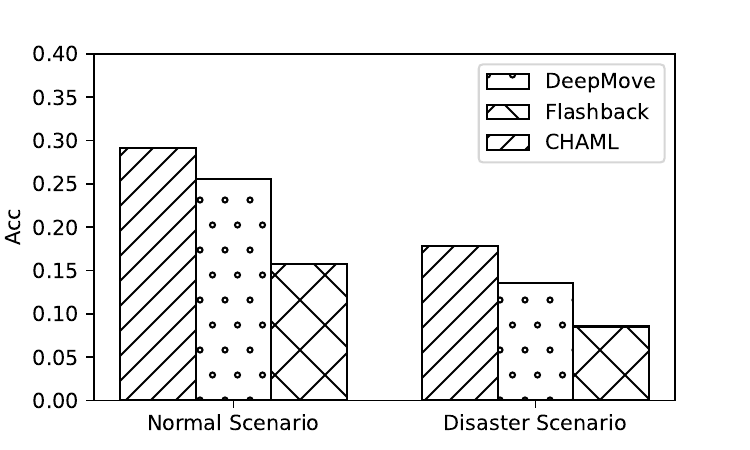}
    \caption{\small Accuracy of mobility prediction in normal/extreme scenarios}
    \label{fig:intro:acc}
  \end{subfigure}
  \hfill  % 在两个子图之间添加水平填充以保持间距
  \begin{subfigure}[b]{0.23\textwidth}
    \includegraphics[width=\textwidth]{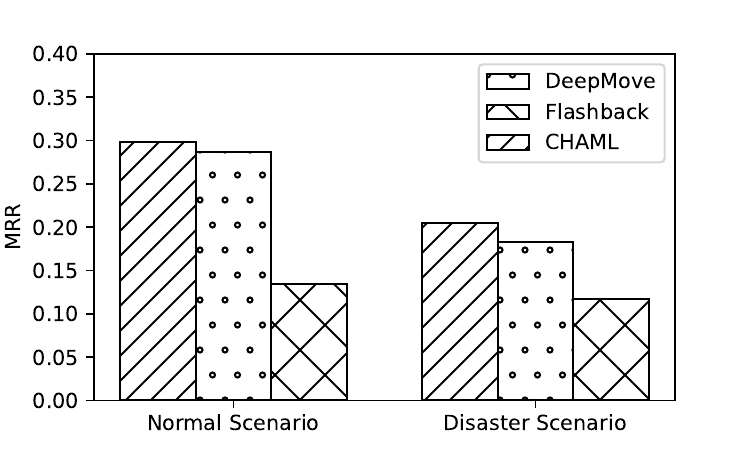}
    \caption{\small Mean reciprocal rank of mobility predictions}
    \label{fig:intro:F1}
  \end{subfigure}

  % 第二行
  \begin{subfigure}[b]{0.23\textwidth}
    \includegraphics[width=\textwidth]{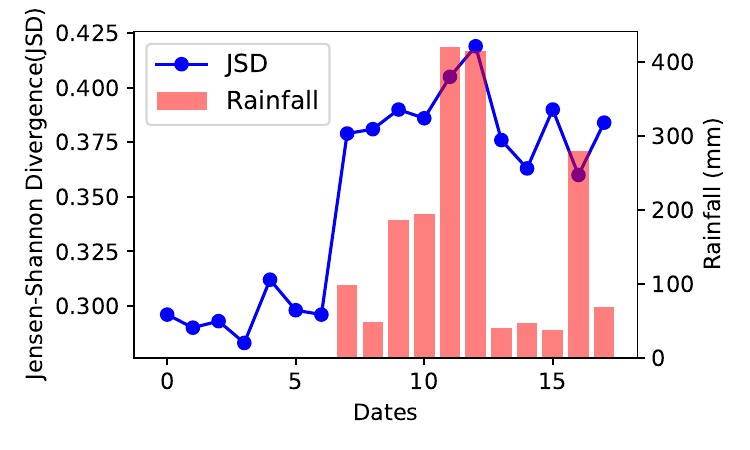}
    % \caption{JSD of location distribution between each day and the average distribution in normal scenarios vs. precipitation of each day}
    \caption{\small JSD between daily location distribution and average location distribution vs. precipitation}
    \label{fig:intro:JSD}
  \end{subfigure}
  \hfill  % 在两个子图之间添加水平填充以保持间距
  \begin{subfigure}[b]{0.23\textwidth}
    \includegraphics[width=\textwidth]{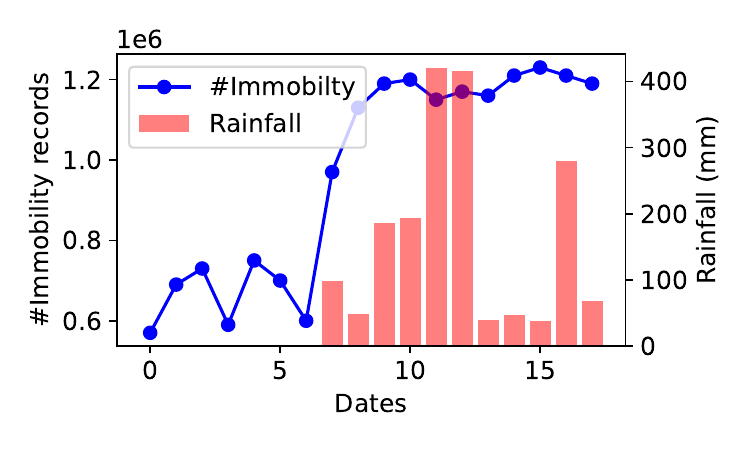}
    \caption{\small The number of immobility records in each day vs. precipitation}
    \label{fig:intro:immob}
  \end{subfigure}
  \caption{Analysis of mobility prediction performance of existing algorithms trained in the normal scenario (i.e., DeepMove and Flashback) as well as cross-city transfer learning algorithm (i.e., CHAML), and the cause of their underperformance.}
  \label{fig:intro}
\end{figure}

As a classic machine learning problem, human mobility prediction has been studied for decades; however, most existing work~\cite{feng2018deepmove,luca2021survey} has focused on normal scenarios rather than extreme scenarios.
As illustrated in Fig.~\ref{fig:intro}(a) and (b), we employ two representative algorithms trained in the normal scenario, i.e., DeepMove~\cite{feng2018deepmove} and Flashback~\cite{yang2020location}, to predict human mobility in normal scenarios and extreme scenarios, respectively. Their performance in extreme scenarios significantly decreases compared with normal scenarios, with an average relative performance gap of 46.4\% and 24.5\% in terms of accuracy and mean reciprocal rank, respectively.
This underperformance arises from the significant difference in mobility patterns between normal and extreme scenarios in terms of spatial distribution (see Fig.~\ref{fig:intro}(c)) and immobility probability~\cite{moro2021mobility} (see Fig.~\ref{fig:intro}(d)).
Therefore, mobility prediction in extreme scenarios remains an open problem.%, that is unsolved by existing mobility prediction algorithms designed for normal scenarios, 

However, developing a specialized mobility prediction model for extreme scenarios is also a challenging task.
The most critical issue lies in the diversity of extreme event situations, e.g., different rainfalls ranging from once-in-a-decade to once-in-a-century, which are usually beyond our experience. It indicates the lack of available mobility data for a specific extreme event in a particular city to train the specialized model before the extreme event~\cite{li2022spatiotemporal}.
A promising solution to this problem is to leverage data from similar extreme events in other cities, transferring knowledge to train the event-specialized model effectively.
However, we also observed that simply utilizing mobility data of similar extreme events from other cities through transfer learning techniques still cannot achieve satisfactory performance. Specifically, the CHAML algorithm~\cite{chen2021curriculum} still exhibits a substantial relative performance gap of 38.9\% and 31.4\% under extreme scenarios compared to normal scenarios in terms of accuracy and mean reciprocal rank (see Fig.~\ref{fig:intro}(a)).
The largely different distribution of spatial venues in source cities and target cities makes it difficult to effectively transfer knowledge. Further, existing cross-city mobility prediction algorithms~\cite{xu2024crosspred,ding2019learning} also fail to utilize the massive trajectory data in normal scenarios of source cities. An ideal approach should effectively utilize trajectory data from the source in both normal and extreme scenarios while decoupling distribution differences of spatial venues to enable efficient transfer of common knowledge describing the shift of mobility patterns (e.g., from the perspective of mobility intention) between normal and extreme scenarios.

In this paper, we propose \textbf{X-MLM}, a cross-e\textbf{X}treme-events \textbf{M}obility-\textbf{L}anguage-\textbf{M}odel with cross-city transfer learning for mobility prediction in extreme scenarios.
%， which also utilize cross-city knowledge and target disaster knowledge at the intention level.
% In this framework, we transfer how disaster affects human mobility knowledge at intention level~\cite{wang2024cola} since where an individual will go in the next timestamp is largely driven by his/her intention~\cite{yuan2014human,yi2018mining,li2024limp}. Furthermore, owing to the fact that LLMs have been proven capable of understanding human behavior~\cite{zhang2024large,zhou2024urban}, we introduce LLMs to transfer the human mobility knowledge between different cities and disasters at the intention level.
In this framework, we leverage LLMs to transfer knowledge about how extreme events influence human mobility at the intention level, facilitating the understanding and prediction of mobility intentions across different cities in extreme scenarios.
Specifically, we propose an RAG-Enhanced Intention Predictior to retrieve similar trajectories with the target city and target extreme event situations as external knowledge. We also propose an intention translator and predictor that aligns intention modality and language modality to enhance LLMs to better forecast the subsequent intention. 
%to predict the next intention through aligning intention modality and language modality to enhance the understanding of LLMs on human intention. 
% In the LLM-based Intention Refiner, we introduce an intention-incorporated prompt based on \textbf{C}hain-\textbf{o}f-\textbf{T}hought (CoT) and a soft prompt to stimulate LLM in understanding human behaviors and how different disaster affects the mobility patterns at intention level to refine the predicted intention. 
In the subsequently introduced LLM-based Intention Refiner, we utilize an intention-incorporated prompt based on Chain-of-Thought (CoT) prompt for reasoning and a event-level-awaring soft prompt
to refine the predicted intentions incorporating the shift of intention from normal to extreme scenarios.
%These elements are designed to enhance the LLM's understanding of human behavior and how different disasters impact mobility patterns at the intention level, thereby refining the predicted intentions."
Finally, we introduce an Intention-Modulated Location Predictior combining an arbitrary base mobility prediction model and intention information to predict the next location of the target trajectory in extreme scenarios.

Our contributions can be summarized as follows:
\begin{packed_itemize}
\item We propose X-MLM, an LLM-based cross-city transfer learning framework, which utilizes LLM-adapted intention embeddings to modulate arbitrary base models to predict human mobility in extreme scenarios. It uses LLM to understand how different extreme events affect human mobility and transfer it across cities and scenarios.
\item We design an Intention-CLIP to achieve the modality alignment between intention and language modality to enhance the ability to understand human mobility intention for LLMs and predict the next intention.
\item Extensive experiments demonstrate that, compared to baselines, X-MLM achieve a 32.8\% improvement in Acc@1, a 28.3\% improvement in MRR, and a 35.0\% improvement in the F1-score for immobility, effectively compensating for the performance decrease in extreme scenarios of base models in normal scenarios.
\end{packed_itemize}

\section{Related Works}
\iffalse
There are many existing works on predicting the next location or POI of the given trajectory~\cite{wang2019urban,luca2021survey}. However, these methods are mainly designed for mobility prediction in normal scenarios, referring to a performance decrease when transferring to a extreme scenario.
\fi

% \para{Deep Learning Mobility Prediction Models.}

\para{Traditional Deep Learning Models in Mobility Prediction}
Since mobility prediction tasks have been widely studied in recent years~\cite{wang2019urban,luca2021survey,zhang2022beyond}, there are many mobility prediction models using deep learning methods. Feng et. al. proposed DeepMove~\cite{feng2018deepmove} to use RNN and attention to extract time and location embeddings for trajectory prediction. Flashback~\cite{yang2020location} enhances the prediction performance of RNNs by emphasizing periodic and contextual information. STiSAN+~\cite{jiang2023spatial} further refines predictions by considering both daily and hourly timing along with spatial relationships. SSDL~\cite{gao2023predicting} represents the complex human mobility semantics in different hidden spaces by decoupling the time-invariant and time-varying factors, and the POI graph structure is designed to explore the heterogeneous cooperation signals in the historical mobility. STAR~\cite{wang2024spatiotemporal} captures spatio-temporal correspondence by designing different spatio-temporal maps and building a stay branch to simulate the stay time in different locations, which is ultimately optimized through adversarial training.

There are also works using transfer learning to extract knowledge from other cities to enhance the prediction in the current city. CHAML~\cite{chen2021curriculum} considers both city-level and user-level difficulty to enhance conditional sampling in the meta-training process. Xu et. al. propose CrossPred~\cite{xu2024crosspred} that takes into account multiple features for cross-city POI matching and uses maximum mean difference (MMD) to enhance shared POI features between cities.

However, these models do not consider the mobility pattern shift in extreme scenarios.
Thus, a series of mobility prediction models specially designed for extreme scenarios are proposed. Song et. al.~\cite{song2014prediction} proposed a system using HMM to model human behavior and use MDP to generate human mobility in extreme scenarios. DeepMob~\cite{song2017deepmob} uses multi-modal learning to combine the information of behavior and mobility by a shared representation in the deep neural network and achieves a performance gain in both behavior and mobility prediction. WS-BiGNN~\cite{xue2024predicting} represents mobile and network search records as link prediction tasks by constructing a bipartite graph, and introduces a time-weighted module and a search-mobile memory module to achieve the prediction of individual irregular mobility.

\para{LLM-enhanced Models in Mobility Prediction} Recently, as LLMs have proved to have great advantages in understanding human behavior, more and more researchers have begun to explore LLMs for mobility prediction~\cite{zhang2024large}. Wang et. al.~\cite{wang2023would} introduced LLMMob to capture both long-term and short-term mobility patterns by considering "historical" and "contextual" stays, and incorporating time information for more accurate predictions. LLM4POI~\cite{li2024large} can understand the intrinsic meaning of context information while maintaining the original format of heterogeneous location data, thus avoiding information loss and realizing POI recommendation. AgentMove~\cite{feng2024agentmove} enhances mobility prediction by mining individual patterns, modeling urban effects, and extracting shared population behaviors. However, these models are designed for mobility prediction in normal scenarios and are unable to handle the mobility shift in extreme scenarios.

\begin{figure*}[t]
  \centering
  \includegraphics[width=0.995\linewidth]{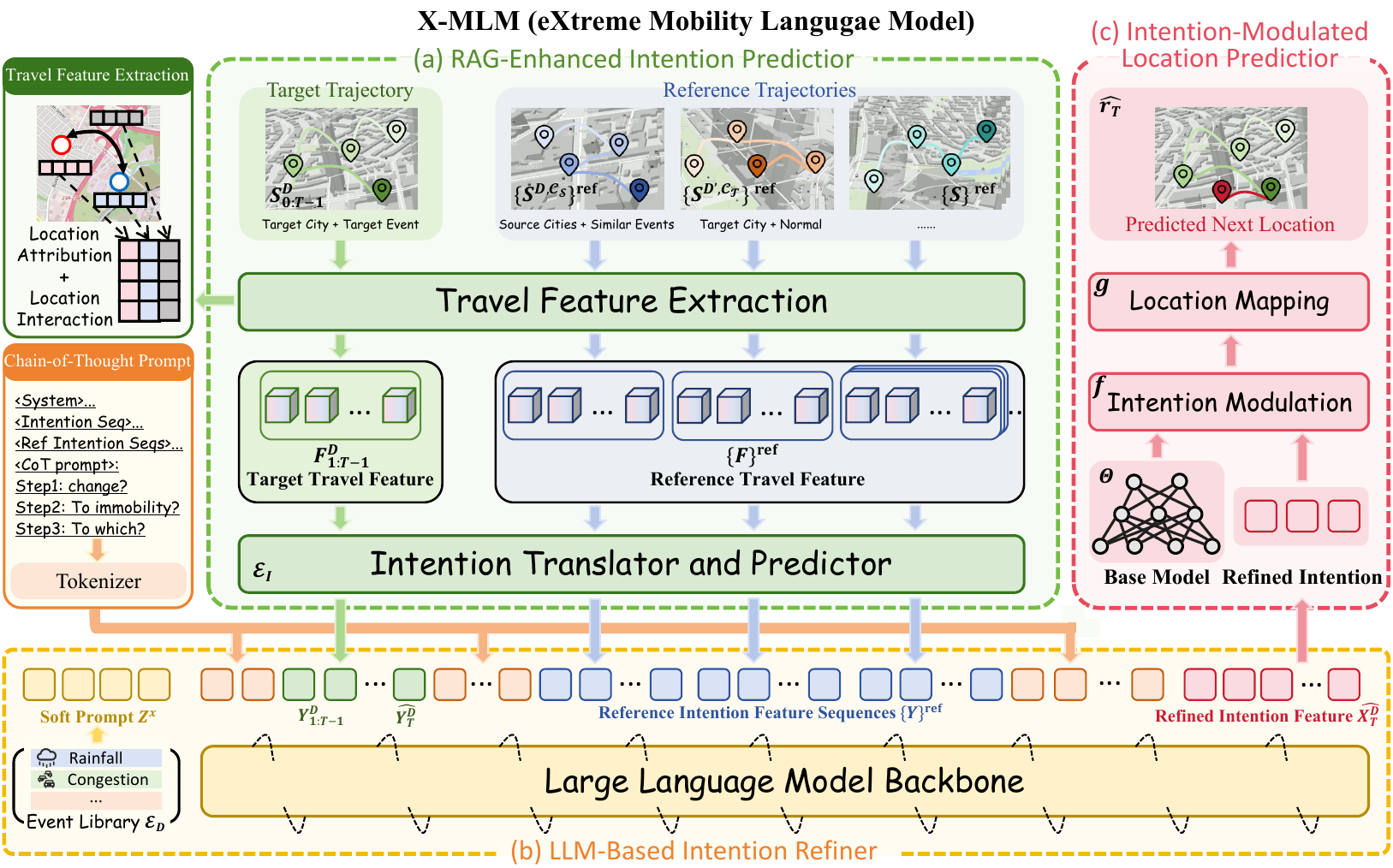}
  \caption{Overall framework of X-MLM}
  \label{fig:main}
\end{figure*}
\section{Problem Definition}
\begin{definition}[Location]
A location $r$ is defined as an area of a specific geographical space such as an administration, a street block, or a community. It is associated with attributes including POI category, POI number, etc. Different locations may also have interactions and relationships based on the road network and transportation facility of the city.
\end{definition}
% In the mobility prediction task, we aim to predict the location where the user will go at the next timestamp given the history trajectory. 
Based on the definition above, the human mobility trajectory can be defined as follows. 
\begin{definition}[Human Mobility Trajectory]
A human mobility trajectory is formulated as a location sequence $S_{0:t}=\{r_0,r_1...r_{t}\}$ in which $r_t$ indicates the location where the individual is located at timestamp $t$. 
\end{definition}

\iffalse
Since data for target extreme events in the target city are sparse, we can use cross-city learning to enhance the prediction performance. We define the target city and source city as follows.
\fi

\iffalse
\begin{definition}[Target City and Source City]
For a specific extreme event, the target city $\mathcal{C}_\mathcal{T}$ refers to the city where the extreme event occurs. Source cities $\mathcal{C}_\mathcal{S}$ refer to all the other cities whose historical trajectories we have access to before the target extreme event.
\end{definition}
\fi

Based on the mobility prediction task~\cite{feng2018deepmove,yang2020location,xue2021mobtcast,feng2024agentmove,moro2021mobility}, we define the task of human mobility prediction in extreme scenarios as follows.
\iffalse
\begin{problem}[Human Mobility Prediction in extreme scenario]
In the extreme scenario in a certain city, given a human mobility trajectory $S_{0:T-1}^{{D}}$ in the extreme scenarios, the mobility trajectory set of this city in normal scenarios $\{S\}^N$ and the extreme event level $d$, the goal is to predict the next location $r_T$ at timestamp $T$. This process can be enhanced by leveraging rich contextual information including information on extreme event level $d$ and a set of reference trajectories $\{S\}^{ref}$.
\end{problem}
\fi
\begin{problem}[Human Mobility Prediction in extreme scenario]
For a given city that experiencing a extreme event, we have a user's mobility trajectory $S_{0:T-1}^{{D}}$ and a set of historical trajectories of this city in normal scenarios $\{S\}^N$, the goal is to predict the next location $r_T$ of the user. This prediction can be enhanced by incorporating external information such as the extreme event level $d$ and mobility trajectories $\{S\}^{ref}$ from other cities in normal and extreme scenarios.
\end{problem}
\section{Method}
\subsection{Overall Framework}
To predict human mobility in extreme scenarios, we propose an LLM-driven transfer learning framework named X-MLM. This framework enhances existing deep mobility prediction models with intention modeled by LLMs and transfers knowledge of how extreme event affects mobility patterns among cities and scenarios.
The deep mobility prediction model used in the framework is referred to as the base model. 
%in the following part. This framework can compensate for the performance decrease in extreme scenarios.
In addition, the city experiencing the extreme event is denoted as the target city, and other cities are denoted as source cities.

%Then we define the city that experiencing the disaster as the target city and other cities as source cities, 
%the information flow of our framework can be illustrated in Fig.~\ref{fig:transfer_overall}.

\begin{figure}[b]
  \centering
  \includegraphics[width=0.95\linewidth]
  {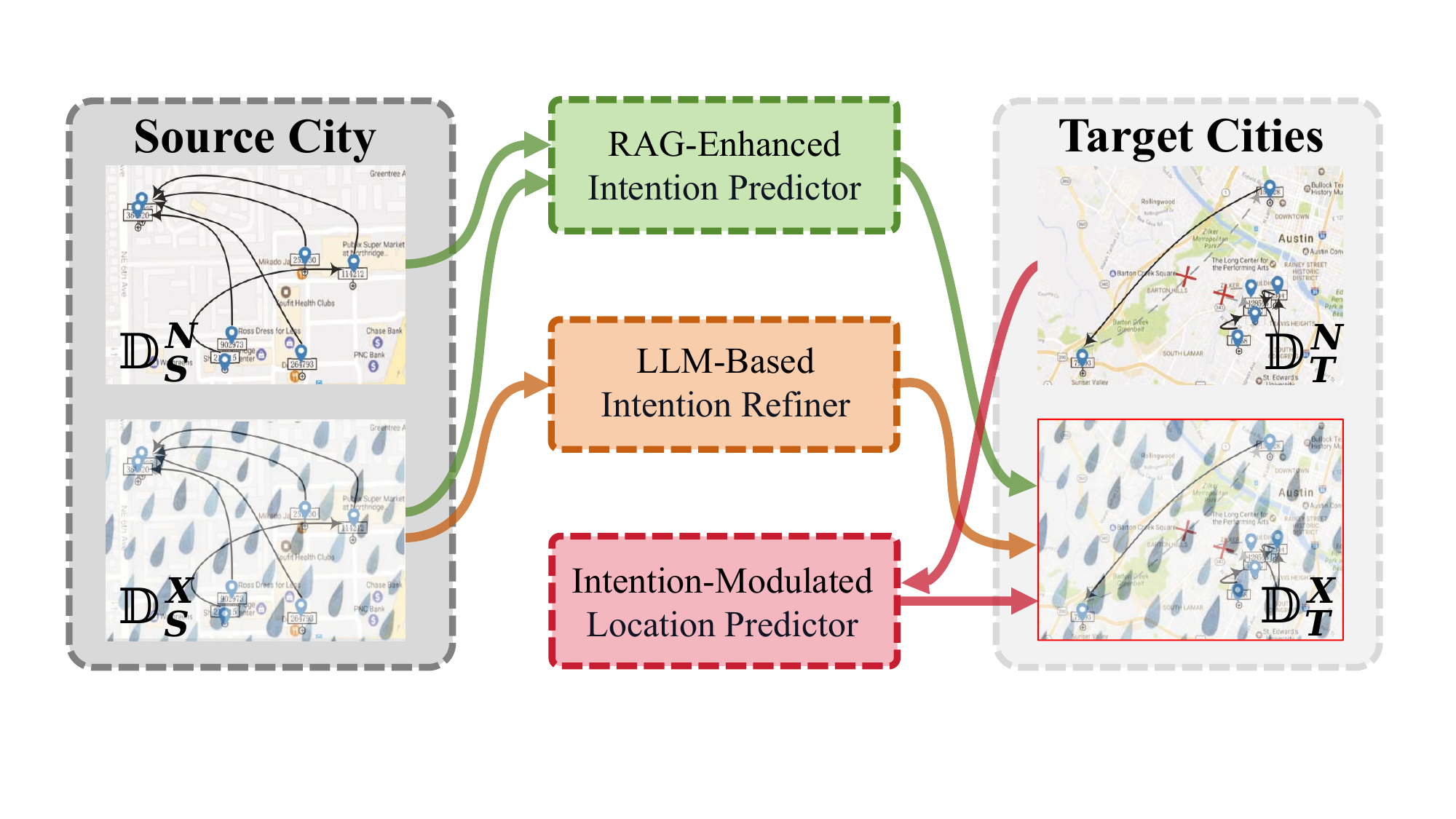}
  \caption{Information flow in our framework, in which $\mathbb{D}^X_S$, $\mathbb{D}^N_S$, $\mathbb{D}^X_T$, $\mathbb{D}^N_T$ refers to 
  the data in extreme scenarios in the source cities, 
  data in normal scenarios in the source cities,
  data in extreme scenarios in the target cities, 
  and data in normal scenarios in the target cities, respectively.}
  \label{fig:transfer_overall}
\end{figure}

The overall structure of our framework is shown in Fig.~\ref{fig:main}, which is composed of three modules with the information flow between them shown in Fig.~\ref{fig:transfer_overall}.
Specifically, we initially introduce an RAG-Enhanced Intention Predictor to extract historical intention sequences from user trajectories with unsupervised learning and predict the next intention.
In this part, we train an intention translator and predictor $\mathcal{E}_\mathbf{I}$ on the data of normal scenarios in both source and target cities as shown in the green rectangle of Intention Prediction in Fig.~\ref{fig:transfer_overall}.
% We use this module to predict the next intention and achieve the modality alignment between intention and language modality.
Then we introduce the LLM-based Intention Refiner to refine the predicted intention.
In this module, we fine-tune an LLM to understand how different extreme events affect human mobility patterns in the data of extreme scenarios of source cities as shown in the orange rectangle of Intention Refinement in Fig.~\ref{fig:transfer_overall}.
Finally, we use an Intention-Modulated Location Predictor with a deep-learning mobility prediction model as the base model to predict the specific location of the next timestamp with intention modulation.
In this part, we train the base model in normal scenarios of the target city to learn how intention affects the prediction of the next location in the target city as shown in the red rectangle of Location Prediction in Fig.~\ref{fig:transfer_overall}.

Thus, the framework can transfer the knowledge of how different extreme event affects human mobility patterns in different cities and predict the next location in extreme scenarios.
\subsection{RAG-Enhanced Intention Predictor}
In this module, we train an intention translator and predictor $\mathcal{E}_\mathbf{I}$ to predict the next intention of the given trajectory. Furthermore, we also retrieve a set of reference trajectories to address the data sparsity problem of trajectories in the target city for the target extreme event.
\subsubsection{Intention Translator and Predictor}
As mentioned in Sec.~\ref{sec:introduction}, the mobility patterns shift in the extreme scenario, and it is difficult to transfer between cities. 
% To address the heterogeneity of mobility patterns between different cities
Thus, we convert the trajectory $S^D_{0:T-1}$ to the intention level and predict the next intention $X^D_T$ using an intention translator and predictor $\mathcal{E}_\mathbf{I}$. 

Since the intention is a concept related to travel between locations, i.e. location switch, as defined in~\cite{yi2018mining,li2024limp}, we convert the trajectory $S^{D}_{0:T-1}$ to a travel feature sequence $F^{D}_{1:T-1}$ through travel feature extraction~\cite{he2020human,yuan2014human,yi2018mining}.
Specifically, given a travel between two locations in a source city, we extract the location attribution of these two locations including the number of each POI category in the location, the closest distance to the nearest transportation facility POI, etc., and interactions between two locations such as the number of each type of road between two locations as the travel feature. Then we map the feature of each travel in the travel feature sequences to a shared intention space to obtain the intention feature sequences in intention modality $X^{D}_{1:T-1}$ for each trajectory. This process, defined as intention mapping, includes a Transfer Component Analysis (TCA)~\cite{pan2010domain} on all travel feature sequences in all source cities and an unsupervised clustering. We consider each cluster as an intention following~\cite{he2020human} and use the feature of the central sample of each cluster as the target intention feature, thus obtaining an intention feature sequence for each trajectory. Aside from that, in the intention feature sequences, we separately assign an immobility embedding to all travels in which the location is not changed. We define the data form obtained from the above method as intention modality.
% Thus, we can obtain the intention feature sequence in the intention modality $X^{D} \in \mathcal{R}^{T \times D_{int}}$, in which $D_{int}$ refers to the dimension of each intention.

However, the LLMs cannot understand these intention features because they are pre-trained in the language modality~\cite{radford2021learning} while the intention feature sequence is in the intention modality. Thus, we use a transformer that receives a travel feature sequence $F^{D}_{1:T-1}$ and outputs the embedding of the next intention $Y^{D}_T$ at language modality to align the intention feature sequence to the language modality. This process can be formulated as follows:
\begin{align}
    Y^{D}_T = \mathcal{E}_\mathbf{I}\left(F^{D}_{1:T-1}\right),
    \label{Eint}
\end{align}
in which $F^{D}_{1:T-1}$ refers to any travel feature sequences including the given trajectories and reference trajectories. It should be noticed that the output is the next intention $Y^D_T$ because we use the time series modeling capability of the transformer to predict the next intention. 
\subsubsection{Intention-CLIP}
\begin{figure}
  \centering
  \includegraphics[width=0.999\linewidth]{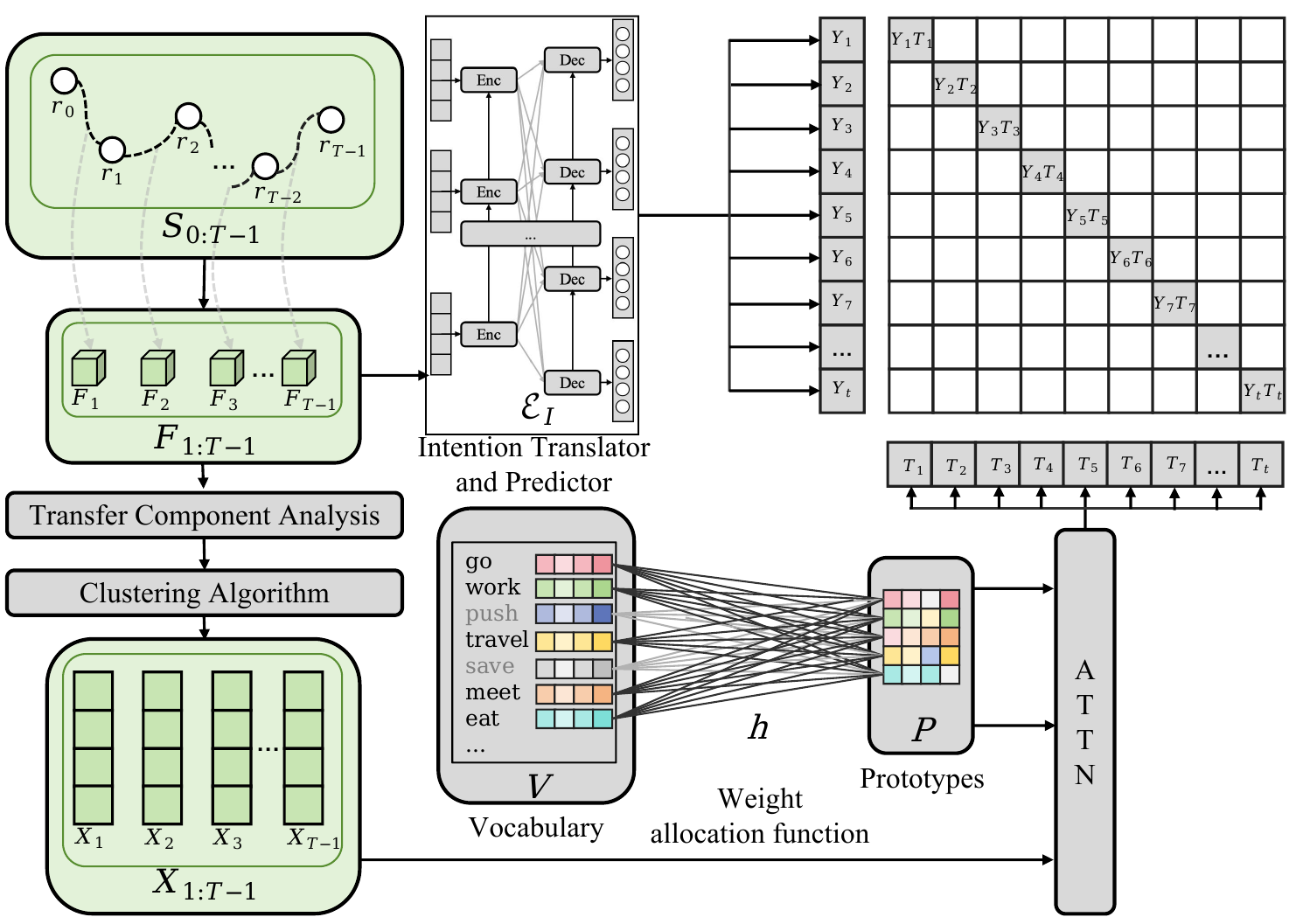}
  \caption{Illustration of Intention-CLIP}
  \label{fig:intention_clip_main}
\end{figure}

To train the intention translator and predictor $\mathcal{E}_\mathbf{I}$, we propose \textbf{C}ontrastive \textbf{L}anguage-\textbf{Intention} \textbf{P}re-training (\textbf{Intention-CLIP}) illustrated in Fig.~\ref{fig:intention_clip_main}. It achieves the prediction of the next intention and the alignment of the language-intention modality at the same time. Inspired by~\cite{jin2023time,misra2023reprogramming,yang2021voice2series}, we design an attention mechanism based on LLM's vocabulary table to achieve alignment. We firstly pass the vocabulary table of the LLM $V \in \mathcal{R}^{N_V \times D_V}$ through a weight allocation function $h$ to obtain the matrix of intention-related prototypes $P \in \mathcal{R}^{N_P \times D_V}$ indicating a compressed intention feature. The weight allocation function $h$ represents the weights of all combinations of the intention-related vocabularies to the prototypes. This process can be formulated as follows:
\begin{align}
    P = h(V),
    \label{proto}
\end{align}
in which $N_V$ refers to the number of LLM vocabulary, $D_V$ refers to the dimension of LLM vocabulary embeddings and $N_P$ refers to the number of prototypes, which is a hyper-parameter.
% Then we use an attention mechanism to obtain the attention of the last intention embedding of the last element of the given intention embedding sequence and all prototypes. Then we pass this embedding to a Linear layer to obtain the intention embedding at the language modality $Y_t$, which can be formulated as:
Then, we use an attention mechanism to obtain the relationship between the intention feature and prototypes by using $X_t$ as query and $P$ as key and value to obtain the intention embedding at the language modality $T_t$.
\begin{align}
    T_t = \text{softmax}\left( \frac{X_tP^T}{\sqrt{d_k}}\right)P.
    \label{attention}
\end{align}

When conducting Intention-CLIP, we obtain the trajectory embedding by feeding the first $T-1$ elements in the location intention-oriented sequence into the intention translator and predictor $\mathcal{E}_\mathbf{I}$ to obtain the prediction of the next intention $Y_t$. We then update the weights of the intention translator and predictor, the attention mechanism, and the weight allocation function by optimizing the InfoNCE loss~\cite{radford2021learning} and cross-entropy loss as follows:
\begin{align}
    \mathcal{L}_{C} = \text{InfoNCE}\left( T_t,Y_t \right) + \text{CrossEntropy}\left(T_t,Y_t \right),
\end{align}
in which $\text{InfoNCE}$ refers to the InfoNCE loss and $\text{CrossEntropy}$ refers to the Cross-entropy loss. In addition, for all immobility travels, we initialize their embedding as the embedding after passing “stay still” to the tokenizer of the LLM and optimize this embedding together using $\mathcal{L}_{C}$. Thus, as $\mathcal{E}_I$ is trained in $\mathbb{D}^{N}_{S}$ and $\mathbb{D}^{N}_{T}$ as mentioned in Fig.~\ref{fig:transfer_overall}, it can learn the intention transfer patterns in normal scenarios and output the feature of the next intention in the language modality $Y_T$ given an intention-oriented feature sequence $X_{0:T-1}$.

% Since we use a carefully designed Intention-CLIP, which will be introduced in Sec~\ref{sec:training}, the intention translator and predictor essentially realizes not only the alignment of the intention modality with the language modality but also the preliminary prediction of the intention at the next timestamp. Thus, we put the given trajectory and the reference trajectories through the intention translator and predictor to get their next intention embeddings, respectively, and use them to construct the prompt of the LLM. The disaster encoder is a simple linear layer that receives a textual feature of the disaster level and outputs an embedding of the current disaster level to address the heterogeneity between different disasters.

\subsubsection{Trajectory Retrieval}

% To address the sparsity of trajectories in the target city of the target disaster, we retrieve reference trajectories set $\{S^{ref}\}$ that are most similar to $S_{0:T-1}^D$ from the database as external knowledge for the prediction. It includes trajectories that are most similar to $S_{0:T-1}^D$ at the intention level among all trajectories with disaster level $d$ in the source cities $\{S^{D,\mathcal{C}_\mathcal{S}}\}^{ref}$, and trajectories that are most similar to the intention feature sequence of $S_{0:T-1}^D$ among all trajectories in the target city in normal scenarios or other extreme scenarios $\{S^{D',\mathcal{C}_\mathcal{T}}\}^{ref}$. 
% We use the travel feature extraction and intention mapping to obtain their intention feature sequences $X^D_{1:T-1}$ and $X'$ and use Dynamic Time Wrapping~\cite{muller2007dynamic} to measure the similarity between two trajectories. The process can be formulated as:
% \begin{align}
%     \{S\}^{ref} = \{S^{D,\mathcal{C}_\mathcal{S}}\}^{ref} \cup \{S^{D',\mathcal{C}_\mathcal{T}}\}^{ref}.
%     \label{ref_traj}
% \end{align}
% A more detailed description of the trajectory retrieval process can be found in Appendix~\ref{sec:retrieval}.

To address the sparsity of trajectories in the target city of the target extreme event, we retrieve reference trajectories that are most similar to the given trajectory $S_{0:T-1}^{D}$ from the trajectory database as external knowledge for the prediction. The set of these reference trajectories is formulated as $\{S\}^{\text{ref}}$.

Specifically, we initially construct a trajectory database, which includes human mobility trajectories for all cities in all extreme scenarios and normal scenarios in history. For the given trajectory $S_{0:T-1}^{D}$ and the extreme event information, we retrieve two categories of trajectories from the database. The first category is trajectories that are most similar to the given trajectory at the intention level among all trajectories during extreme events in the source cities, which is formulated as $\{S^{D,\mathcal{C}_\mathcal{S}}\}^{\text{ref}}$. This approach addresses the data sparsity issue for the target extreme event, as intuitively, even if the target extreme event hasn't occurred in the target city, it is much more likely to have occurred in one of the source cities.
% We therefore retrieve all trajectories that are similar to those under the target disaster to assist the prediction. 
The second category is the set of trajectories that are most similar to the given trajectory's intention feature sequence among all trajectories in the target city in normal scenarios or other extreme scenarios, which is formulated as $\{S^{D',\mathcal{C}_\mathcal{T}}\}^{\text{ref}}$. Thus, we address the heterogeneity of mobility between different cities. This is due to the heterogeneity of mobility patterns across different cities, which indicates that the knowledge of mobility obtained from other cities is not necessarily fully applicable to the target city.

Specifically, for the given trajectory $S^D_{0:T-1}$ and each trajectory $S'$ in the database, we use the travel feature extraction and intention mapping to obtain their intention feature sequences $X^D_{1:T-1}$ and $X'$. Then we calculate the similarity between these two sequences using Dynamic Time Wrapping~\cite{muller2007dynamic} and select the set of the top $k$ most similar trajectories in the above two categories. Then we can get the set of reference trajectories as follows:
\begin{align}
    \{S\}^{\text{ref}} = \{S^{D,\mathcal{C}_\mathcal{S}}\}^{\text{ref}} \cup \{S^{D',\mathcal{C}_\mathcal{T}}\}^{\text{ref}}.
    \label{ref_traj}
\end{align}

\subsection{LLM-based Intention Refiner}\label{sec:intention_prediction_and_modality_alignment_module}

Although we have initially predicted the feature of the next intention $\hat{Y^D_T}$ in the RAG-Enhanced Intention Predictor, it only captures the common patterns of human mobility in normal scenarios because $\mathcal{E}_\mathbf{I}$ is trained on the dataset for normal scenarios in source cities. Since LLM is proven to be powerful in understanding human behaviors, we use it to refine the predicted next intention feature $Y^D_T$ based on the reference intention feature sequences $\{Y\}^{\text{ref}}$ and the target extreme event level $d$ to get the refined intention feature $\hat{X^D_T}$. 
% Specifically, we introduce a hybrid prompting strategy including an intention-incorporated prompt and a soft prompt~\cite{qin2021learning}. Then we construct a hybrid tokenize strategy to get intention-incorporated embeddings $Z^h$ from the hard prompt and disaster embedding $Z^d$ from the soft prompt to guide the LLM in refining the intention.
Specifically, we introduce a hybrid prompting strategy including an intention-incorporated prompt with immobility modeling and a event-level-awaring soft prompt~\cite{qin2021learning} to stimulate the understanding of human behaviors of LLM in refining the intention.

\subsubsection{Intention-Incorporated Prompt}
Different from the traditional prompting strategy of LLMs which inputs the tokenized embedding of the text instruction, we design an intention-incorporated prompt inspired by~\cite{tang2024higpt,ye2023natural,tang2024graphgpt} to better inspire the LLMs to understand intention information. In this prompt, we encode the text instructions into token embeddings using the tokenizer of the LLM while directly using the intention feature sequence as the embeddings for intention tokens to construct the intention-incorporated embeddings.
% the the part of the prompt that would have used text to input the intention sequence is replaced directly with the intention feature embedding sequence. The text prompt is encoded using the text encoder of the LLM, while the intention sequence does not go through the text encoder but directly uses its intention embedding sequence as the input to the LLM.

\begin{table*}[]
\small
\setlength{\tabcolsep}{2.5pt}  % Reduce column separation
\caption{Structure of the intention-incorporated prompt.}
\label{tab:prompt}
\begin{tabular}{>{\raggedright\arraybackslash}p{18cm}} % Define the width of the column and text alignment
\toprule
\textbf{System Prompt:}\newline You are now a discriminator of predicted human intentions in extreme scenarios, tasked with determining whether the given user's possible next intention is right based on the user's previous intention sequence, extreme event level, a possible next intention and other reference intention sequences. Note that: \newline 1. The intentions are token embeddings, that contain the intention information. It is wrapped in the "``", which means that any token in two "`" refers to the intention embedding rather than the text token. \newline 2. Event level is divided into: \newline  (1)"no extreme event": Indicates that there is no effect on human mobility. \newline (2)"minor extreme event": An individual's daily travel plans may not be affected much, but alternatives should be considered. \newline (3)"general extreme event": Human mobility may be affected by extreme events, causing certain activities to be adjusted or canceled for safety reasons. \newline  (4)"severe extreme event": Human movement will be greatly affected by the extreme event, and the probability of staying still will increase. \newline  3. Reference intention sequences are selected sequences from an intention sequence RAG. They are the most similar sequences to the given intention sequence. You can refer to these sequences to generate your answers but don't copy them exactly. \newline  4. The given possible intention is not necessarily accurate, you need to judge whether it needs to change. \newline 5. Both the user's previous intention sequence and the other reference intention sequences are in the format of a list like [`intention embedding 1`, `intention embedding 2`...].\\ \midrule

\textbf{Intention Sequence Prompt:}\newline Event Level: \textless{}extreme event level\textgreater{}. \newline Intention embedding sequence: \textcolor{blue}{\textless{}representation of the intention sequence\textgreater{}}. \newline The given possible next intention embedding for this sequence is \textcolor{blue}{\textless{}representation of the predicted next intention\textgreater{}}.\\ \midrule
\textbf{Reference Intention Sequence Prompt:} \newline You need to refer to the following sequence to distinguish whether the given possible next intention embedding is right: \newline  Reference Sequence 1: \newline  Event Level: \textless{}extreme event level\textgreater{}. \newline  Intention embedding sequence: \textcolor{blue}{\textless{}representation of the intention sequence\textgreater{}.} \newline The given possible next intention embedding for this sequence is \textcolor{blue}{\textless{}representation of the predicted next intention\textgreater{}}.\newline ... \newline Reference Sequence \textless{}k\textgreater{}: \newline  Event Level: \textless{}extreme event level\textgreater{}. \newline Intention embedding sequence: \textcolor{blue}{\textless{}representation of the intention sequence\textgreater{}}. \newline  The given possible next intention embedding for this sequence is \textcolor{blue}{\textless{}representation of the predicted next intention\textgreater{}}. \\ \midrule
\textbf{CoT Prompt:} \newline Let's think step by step. You need to answer each of the three questions below, and if the answer to the first question is "yes", the following questions will be output as "None": \newline 
(1) Is the given possible next intent embedding right? \newline 
(2) If the answer to the previous question is "yes", this answer is set to "None". If the answer to the previous question is "no", please answer: Given the current event level, should the next intention be "stay still"? \newline 
(3) If the answer to the previous question is "yes", this answer is set to "stay still". If the answer to the previous question is "no", you need to give the index of the correct next intention embedding. \newline 
The indexes and embeddings of the intentions you can choose from are \textless{}intention representation of all intentions\textgreater{}.\\ \midrule
\textbf{Answer Prompt:} \newline Now give your answer. You should output the answer in a ["no", "no", "2"] format, nothing else. \newline Your answer:
\\
\bottomrule
\end{tabular}
\end{table*}

Furthermore, to stimulate the LLM's ability to understand human behavior and to better guide LLMs to understand immobility in the extreme scenario at the intention level, we introduced a Chain-of-Thought prompt framework~\cite{wei2022chain} to refine the intention. The thinking process contains three steps. Firstly, we guide the LLM to decide whether the predicted next intention result is right. If not, we guide LLM to distinguish whether the next intention is immobility. If not, we guide LLMs to decide which intention should be selected as the next intention given all the non-immobility intention embeddings. 
% See Appendix~\ref{app:prompt} for more details.
A complete prompt is shown in Table~\ref{tab:prompt}.

\subsubsection{Event-Level-Awaring Soft Prompt}
In order to address the heterogeneity of mobility patterns between different extreme events, we use an Event Library $\mathcal{E}_\mathbf{D}$ to construct an adaptive embedding $Z^x$ given the event information to model the extreme event level. The event library is instantiated as a look-up table. We use the event embedding $Z^x$ in a soft prompt paradigm to model the heterogeneity of different extreme events. Different from the traditional hard prompt, a soft prompt adjusts the behaviors of the LLM by learning a set of trainable embeddings to be used as prefixes for the input sequences. This approach can significantly reduce the number of parameters that need to be updated while maintaining or improving the performance of the model. 
% Specifically, we use soft prompts to learn unique embedding representations for each different disaster level to model how they affect human mobility patterns.
% Specifically, we dynamically optimize the disaster embedding in the fine-tuning process to learn how different disaster affects human mobility patterns.
Thus, the construction process of the input embedding sequence is the combination of a soft prompt and the text embedding sequences with the intention-incorporated embeddings.

When fine-tuning the LLM to learn how different extreme event affects human mobility patterns, we use prefix-tuning~\cite{li2021prefix} to fine-tune the LLM without changing most of the parameters. We build the input embedding sequence and input it into the backbone of LLM without the tokenizer to obtain the predicted next intention label. Then, we use it to index the corresponding intention feature in intention modality to obtain the feature for the next intention $\hat{X_T^{D}}$ for the given trajectory in the target city. Then we use cross-entropy loss to optimize the model. The loss function can be formulated as follows:
\begin{align}
    \mathcal{L}_P = \text{CrossEntropy}(\hat{X_T^{D}},X_T^{D}).
\end{align}
\subsection{Intention-Modulated Location Predictior}
After obtaining the refined intention feature $\hat{X_{T}^{D}}$, we use the Intention-Modulated Location Predictor to predict the specific location where the individual will go in the next timestamp $\hat{r_T}$ with intention modulation. As we highlighted in Sec.~\ref{sec:introduction}, existing models have achieved relatively accurate human mobility prediction in normal scenarios, and intentions can help us to better transfer the prediction in different cities and different scenarios. 

Existing deep learning human mobility prediction models can be considered as a multi-classification task for all candidate locations. Their model structure can be divided into a mobility embedding extraction network $\textbf{H} = \Theta(S)$ to extract embedding for location and the last classification layer to map the mobility embedding to the location $\hat{r_t} = \text{softmax}\left(\text{MLP}(\textbf{H})\right)$, in which $\Theta$ refers to the mobility prediction model without the last classification layer, $\textbf{H} \in \mathcal{R}^{D_H \times 1}$ refers to the embedding for the classification task and $D_H$ refers to the hidden dimension.
% \begin{align}
%     \textbf{H} &= \Theta(S),\\
%     \hat{r_t} &= \text{softmax}\left(\text{MLP}(\textbf{H})\right),
%     \label{NNPM}
% \end{align}
We use an intention modulation function $f$ to incorporate the intention $X$ into existing human mobility prediction models to predict the location. This function can be instantiated into three operations including element-wise production, concatenation, and attention mechanism. The process can be formulated as follows:
\begin{align}
\hat{r_T} = \text{softmax}\left(g\left(f(\textbf{H}, X)\right)\right), \label{intention_modulation}
\end{align}
% \begin{align}
% \text{Element-wise Product: } \hat{r_T} = \text{softmax}\left(g(\textbf{H} \circ X)\right), \\ 
% \text{Concatenation: } \hat{r_T} = \text{softmax}\left(g(\textbf{Cat}(\textbf{H},X))\right), \\ 
% \text{Attention: } \hat{r_T} = \text{softmax}\left(g(\textbf{Attn}(\textbf{H},X))\right).
%     \label{3ways}
% \end{align}
in which $g$ refers to the location mapping function and $\hat{r_T}$ refers to the prediction of the next location. 
% See Appendix~\ref{sec:intention_modulation} for a more detailed description of the intention modulation function.
We use the intention modulation function $g$ to incorporate intention $X$ into existing human mobility prediction models. The intention modulation function can be instantiated into three operations including element-wise production, concatenation, and attention mechanism as follows:
\begin{align}
\text{Element-wise Product: } \hat{r_T} = \text{softmax}\left(g(\textbf{H} \circ X)\right), \\ 
\text{Concatenation: } \hat{r_T} = \text{softmax}\left(g(\textbf{Cat}(\textbf{H},X))\right), \\ 
\text{Attention: } \hat{r_T} = \text{softmax}\left(g(\textbf{Attn}(\textbf{H},X))\right).
    \label{3ways}
\end{align}

When training the Intention-Modulated Location Predictor, we compute the cross-entropy loss $\mathcal{L}_M$ and use it to optimize the model, which can be formulated as follows:
\begin{align}
    \mathcal{L}_M = \text{CrossEntropy}(\hat{r_T},r_T),
\end{align}
in which $r_T$ refers to the ground truth next location of the given trajectory.

\section{Experiments}
\subsection{Datasets}\label{sec:dataset}

% Through collaboration with a popular mobile application vendor,
% we collect 7 real-world human trajectory datasets in 7 cities during periods of natural disasters, whose statistics are presented in Appendix~\ref{sec:dataset}.
Through collaboration with a popular mobile application vendor,
we collect 7 real-world human trajectory datasets in 7 cities during periods of natural extreme events.
Table~\ref{tab:dataset} shows the statistics of our dataset in which $D_{n}$ and $D_{x}$ refer to the number of days in normal and extreme scenarios, respectively.
In addition, $\#S_{n}$ and $\#S_{x}$ are the number of trajectories in normal and extreme scenarios, and $M_p$ is the maximum precipitation during the extreme event.
Note that the precipitation data is collected from the CHIRPS dataset~\cite{funk2015climate} and
there are five categorized event levels of rainfall.

In particular, each dataset contains historical requested positioning data from mobile users,
exhibiting spatial and temporal characteristics of human mobility in extreme scenarios.
We split them by setting Zhongshan as the target city and the other cities as source cities.
%The statistics of the dataset are in Appendix~\ref{sec:dataset}.

% In our experiment, we collect trajectories in seven cities from a popular mobile application vendor.
% It captures the spatial and temporal data of mobile users each time they request location services within applications, such as when checking in or using location-based social networks.
% It the collection duration, there is at least one disaster in each city.
% For dataset splitting, we choose Zhongshan as the target city and the others as source cities. The basic statics of the data are shown in Table~\ref{tab:dataset}. The rainfall data are 
\subsection{Baselines and Experimental Settings}
\para{Baselines}
Different categories of baselines are compared with X-MLM. Specifically, we select traditional mobility prediction models including \textbf{LSTM}~\cite{hochreiter1997long}, \textbf{GRU}~\cite{cho2014learning}, \textbf{DeepMove}~\cite{feng2018deepmove}, \textbf{Flashback}~\cite{yang2020location}, and \textbf{STiSAN+}~\cite{jiang2023spatial}.
We also select \textbf{HMM+MDP}~\cite{song2014prediction} and \textbf{DeepMob}~\cite{song2017deepmob} which are specially designed for extreme scenarios.
Furthermore, we select models using cross-city knowledge to enhance the prediction performance including \textbf{CHAML}~\cite{chen2021curriculum} and \textbf{CATUS}~\cite{sun2023pre}. Except for traditional deep mobility prediction models, we also use mobility prediction models enhanced by LLM, such as \textbf{LLM4POI}~\cite{li2024large} and \textbf{ST-MoE-BERT}~\cite{he2024st}. 
% A more detailed description of baselines can be found in Appendix~\ref{sec:baseline}.
A more detailed description of baselines are as follows:
% We use following baselines to evaluate the performance of our X-MLM:
\begin{itemize}
    \item \textbf{LSTM}~\cite{hochreiter1997long} uses cell states and gating mechanisms to address gradient issues in traditional RNNs.
    \item \textbf{GRU}~\cite{cho2014learning} learns long-term dependencies and reduces complexity by merging gates and using hidden states as cell states. 
    \item  \textbf{DeepMove}~\cite{feng2018deepmove} combines RNNs and attention mechanisms to predict human mobility by capturing spatial-temporal patterns.
    \item  \textbf{Flashback}~\cite{yang2020location} integrates long-term history data and short-term behavioral patterns, emphasizing periodic and contextual information, to accurately predict users' next location. 
    \item \textbf{STiSAN+}~\cite{jiang2023spatial} captures sequential dependencies by assigning attention weights among different elements, improving the performance in location services for handling spatial-temporal factors and modeling individual behavior patterns. The methods above are designed for normal scenarios, which lack of special design for extreme scenarios. There are also some fundamental models which are designed for mobility prediction in extreme scenarios.
    \item \textbf{HMM+MDP}~\cite{song2014prediction} combines HMM and MDP to simultaneously predict behaviors and mobility paths of individuals after the occurrence of extreme events. 
    \item \textbf{DeepMob}~\cite{song2017deepmob} enables collaborative prediction of both features by jointly modeling behavior and mobility. However, their performance is restricted by limited knowledge of human mobility across different cities. 
    \item \textbf{CHAML}~\cite{chen2021curriculum} is a framework for cross-city POI search recommendations that tackles data scarcity and sample diversity in cold-start cities by integrating difficult-aware meta-learning and city-level learning strategies. 
    \item \textbf{CATUS}~\cite{sun2023pre} is a pre-trained model that learns common transition patterns across cities using self-supervised tasks, enhanced with class transition samplers and implicit-explicit policies. However, these methods fail to account for extreme scenarios.
    \item \textbf{LLM4POI}~\cite{li2024large} predicts human mobility by converting location social network data into Q\&A format and combines historical tracks with key query similarity to leverage contextual information. 
    \item \textbf{ST-MoE-BERT}~\cite{he2024st} is a spatial-temporal framework that integrates BERT models and a Mixture of Experts (MoE) architecture for predicting long-term human mobility patterns across cities. While they make use of the capability of LLMs to comprehend human behaviors, they are not originally designed for extreme scenarios.
\end{itemize}

\begin{table}[t]
\centering
\caption{The statistical information of the datasets.}
\small
\renewcommand{\arraystretch}{1.2} % 增加行高，使表格更易读
\begin{tabularx}{0.5\textwidth}{l|l|>{\centering\arraybackslash}p{0.25cm}>{\centering\arraybackslash}p{0.25cm}>{\centering\arraybackslash}p{1.2cm}>{\centering\arraybackslash}p{1.2cm}>{\centering\arraybackslash}p{0.85cm}}
\hline
\textbf{Usage} & \textbf{Dataset} & \textbf{$D_{n}$} & \textbf{$D_{x}$} & \textbf{$\#S_{n}$} & \textbf{$\#S_{x}$} & $M_{p}$ \\ \hline
               & Qingyuan         & 7                    & 14                     & 4,858,142              & 8,000,537      &127.08          \\
               & Shaoguan         & 7                    & 14                     & 2,812,149              & 4,764,873     & 97.04           \\
               & Zhuhai           & 7                    & 6                      & 3,113,070              & 6,558,081       & 75.46         \\
               & Wuzhou           & 7                    & 8                      & 2,609,726              & 2,448,882      & 96.43         \\
               & Guilin           & 6                    & 8                      & 9,733,964              & 15,686,556       & 83.39        \\
\multirow{-6}{*}{Source} & Hezhou           & 7                    & 14                     & 4,568,532              & 6,124,276      & 88.48          \\ \hline
Target & Zhongshan        & 7                    & 13                     & 16,227,196             & 30,816,592        &   90.08     \\ \hline
\end{tabularx}
\label{tab:dataset}
\end{table}

\begin{table*}[t]
\centering
\caption{Predictive performance of X-MLM constructed upon different base models in different scenarios, with relative performance differences in terms of different indicators compared to the normal scenarios shown in parentheses.}
\fontsize{4pt}{5pt}\selectfont
\setlength{\tabcolsep}{3pt}  % 减小列间距
\setlength{\arrayrulewidth}{0.1pt}  % 设置线条粗细为 0.3pt
\resizebox{0.99\textwidth}{!}
{
\begin{tabular}{l|ccc|ccc|ccc|ccc|ccc}
\hline
 & \multicolumn{3}{c|}{\textbf{Normal Scenario}} & \multicolumn{3}{c|}{\textbf{extreme scenario}} & \multicolumn{9}{c}{\textbf{extreme scenario with Intention Modulation}} \\
\cline{8-16} 
 & \multicolumn{3}{c|}{} & \multicolumn{3}{c|}{} & \multicolumn{3}{c|}{\textbf{MUL}} & \multicolumn{3}{c|}{\textbf{CONCAT}} & \multicolumn{3}{c}{\textbf{ATTN}} \\
\hline
Base Model & Acc@1 & Acc@10 & MRR & Acc@1 & Acc@10 & MRR & Acc@1 & Acc@10 & MRR & Acc@1 & Acc@10 & MRR & Acc@1 & Acc@10 & MRR \\
\hline
% RNN~\cite{jeffrey1990finding} & \makecell[tc]{0.0607\\(- - -)} & \makecell[tc]{0.1423\\(- - -)} & \makecell[tc]{0.0851\\(- - -)} & \makecell[tc]{0.0108\\(-82.2\%)} & \makecell[tc]{0.0729\\(-42.8\%)} & \makecell[tc]{0.0328\\(-61.4\%)} & \makecell[tc]{0.0521\\(-14.2\%)} & \makecell[tc]{0.1174\\(-17.5\%)} & \makecell[tc]{0.0792\\(-6.9\%)} & \makecell[tc]{0.0513\\(-15.5\%)} & \makecell[tc]{0.1275\\(-10.4\%)} & \makecell[tc]{0.0666\\(-21.7\%)} & \makecell[tc]{0.0689\\(+13.5\%)} & \makecell[tc]{0.1599\\(+12.4\%)} & \makecell[tc]{0.0894\\(+5.0\%)} \\    \hline
LSTM & \makecell[tc]{0.0607\\(- - -)} & \makecell[tc]{0.1527\\(- - -)} & \makecell[tc]{0.0820\\(- - -)} & \makecell[tc]{0.0181\\(-70.2\%)} & \makecell[tc]{0.0892\\(-41.6\%)} & \makecell[tc]{0.0436\\(-46.8\%)} & \makecell[tc]{0.0581\\(-4.3\%)} & \makecell[tc]{0.1293\\(-15.3\%)} & \makecell[tc]{0.0651\\(-20.6\%)} & \makecell[tc]{0.0626\\(+3.1\%)} & \makecell[tc]{0.1688\\(+10.5\%)} & \makecell[tc]{0.0879\\(+7.2\%)} & \makecell[tc]{0.0640\\(+5.4\%)} & \makecell[tc]{0.1597\\(+4.6\%)} & \makecell[tc]{0.0884\\(+7.8\%)} \\   \hline
GRU & \makecell[tc]{0.0307\\(- - -)} & \makecell[tc]{0.1527\\(- - -)} & \makecell[tc]{0.0786\\(- - -)} & \makecell[tc]{0.0167\\(-45.6\%)} & \makecell[tc]{0.0649\\(-57.4\%)} & \makecell[tc]{0.0359\\(-54.3\%)} & \makecell[tc]{0.0226\\(-26.3\%)} & \makecell[tc]{0.0975\\(-36.1\%)} & \makecell[tc]{0.0601\\(-23.5\%)} & \makecell[tc]{0.0241\\(-21.4\%)} & \makecell[tc]{0.0953\\(-37.5\%)} & \makecell[tc]{0.0575\\(-26.8\%)} & \makecell[tc]{0.0313\\(+1.9\%)} & \makecell[tc]{0.1595\\(+4.4\%)} & \makecell[tc]{0.0738\\(-6.1\%)} \\    \hline
DeepMove & \makecell[tc]{0.2552\\(- - -)} & \makecell[tc]{0.4902\\(- - -)} & \makecell[tc]{0.2864\\(- - -)} & \makecell[tc]{0.1354\\(-46.9\%)} & \makecell[tc]{0.2796\\(-43.0\%)} & \makecell[tc]{0.1832\\(-36.0\%)} & \makecell[tc]{0.2523\\(-1.1\%)} & \makecell[tc]{0.4018\\(-18.0\%)} & \makecell[tc]{0.2695\\(-5.9\%)} & \makecell[tc]{0.2141\\(-16.1\%)} & \makecell[tc]{0.4537\\(-7.4\%)} & \makecell[tc]{0.2294\\(-19.9\%)} & \makecell[tc]{0.2587\\(+1.3\%)} & \makecell[tc]{0.4982\\(+1.6\%)} & \makecell[tc]{0.2905\\(+1.4\%)} \\  \hline
Flashback & \makecell[tc]{0.1576\\(- - -)} & \makecell[tc]{0.3287\\(- - -)} & \makecell[tc]{0.1342\\(- - -)} & \makecell[tc]{0.0854\\(-45.8\%)} & \makecell[tc]{0.1143\\(-65.2\%)} & \makecell[tc]{0.1167\\(-13.0\%)} & \makecell[tc]{0.1534\\(-2.0\%)} & \makecell[tc]{0.3154\\(-4.1\%)} & \makecell[tc]{0.1354\\(+0.8\%)} & \makecell[tc]{0.1521\\(-3.4\%)} & \makecell[tc]{0.3129\\(-5.0\%)} & \makecell[tc]{0.1333\\(-0.6\%)} & \makecell[tc]{0.1579\\(-0.1\%)} & \makecell[tc]{0.3342\\(+1.6\%)} & \makecell[tc]{0.1398\\(+4.1\%)} \\ \hline
STiSAN & \makecell[tc]{0.2891\\(- - -)} & \makecell[tc]{0.5107\\(- - -)} & \makecell[tc]{0.3198\\(- - -)} & \makecell[tc]{0.1872\\(-35.2\%)} & \makecell[tc]{0.3198\\(-37.3\%)} & \makecell[tc]{0.2187\\(-31.6\%)} & \makecell[tc]{0.2761\\(-4.4\%)} & \makecell[tc]{0.4909\\(-3.8\%)} & \makecell[tc]{0.2981\\(-6.7\%)} & \makecell[tc]{0.2799\\(-3.1\%)} & \makecell[tc]{0.4981\\(-2.4\%)} & \makecell[tc]{0.3004\\(-6.0\%)} & \makecell[tc]{0.2897\\(+0.2\%)} & \makecell[tc]{0.5118\\(+0.2\%)} & \makecell[tc]{0.3209\\(+0.3\%)} \\
\hline
\end{tabular}
}

\label{tab:intention_modulation}
\end{table*}

\para{Metrics}
% To compare the performance of X-MLM and all types of baseline methods, we choose average Accuracy@N (i.e., Acc@N and $N=1, 10$), Mean Reciprocal Rank (MRR), and Normalized Discounted Cumulative Gain (NDCG) as general metrics in evaluations.
% Moreover, to further delve into the performance enhancement for immobility, we use Precision (Pre@Immob), Recall (Rec@Immob), and F1-score (F1@Immob) for experimental studies with labeled immobility timestamps. A more detailed introduction of the metrics can be found in Appendix~\ref{sec:metrics}.
In our evaluation process, we choose different metrics to evaluate how different models perform in predicting human mobility in extreme scenarios. Specifically, we choose two types of metrics to evaluate the performance. We use Average Accuracy (Acc@N), Mean Reciprocal Rank (MRR) and Normalized Discounted Cumulative Gain (NDCG) to evaluate the overall performance of all locations, which can be formulated as follows:
\begin{align}
    \text{Acc@N} = \frac{1}{Q} \sum_{q=1}^{Q} \mathbb{I}(\text{correct answer in top } N), \label{Acc@N}
\end{align}
\begin{align}
    \text{MRR} = \frac{1}{Q} \sum_{q=1}^{Q} \frac{1}{\text{rank}_q}, \label{MRR}
\end{align}
in which $Q$ refers to the number of samples and $\mathbb{I}$ is an indication function that returns 1 if the condition is true and 0 otherwise, and $\text{rank}_q$ refers to the rank of the first correct answer in the query $q$. Furthermore, the NDCG can be defined as the division of Discounted Cumulative Gain (DCG) and Ideal Discounted Cumulative Gain (IDCG), which can be formulated as follows:
\begin{align}
    \text{NDCG} = \frac{\text{DCG}}{\text{IDCG}}, \label{NDCG}
\end{align}
in which DCG and IDCG can be formulated as:
\begin{align}
    \text{DCG} = \sum_{i=1}^{n} \frac{2^{rel_i}-1}{\log_2(i+1)}, \label{DCG}
\end{align}
\begin{align}
    \text{IDCG} = \sum_{i=1}^{|R|} \frac{2^{rel_{(i)}}-1}{\log_2(i+1)}, \label{IDCG}
\end{align}
in which $R$ is a list of results sorted by true relevance.

Furthermore, we also use Precision (Pre@Immob), Recall (Rec@Immob), and F1-score (F1@Immob) with respect to immobility timestamps to evaluate the prediction performance in immobility. Specifically, the metrics can be formulated as:
\begin{align}
    \text{Precision} = \frac{\text{TP}}{\text{TP + FP}}, \label{Pre}
\end{align}
\begin{align}
    \text{Recall} = \frac{\text{TP}}{\text{TP + FN}}, \label{recall}
\end{align}
\begin{align}
    \text{F1-score} = 2 \cdot \frac{\text{Precision} \cdot \text{Recall}}{\text{Precision} + \text{Recall}}, \label{F1}
\end{align}
in which $\text{TP}$, $\text{FP}$, and $\text{FN}$ refer to True Positive, False Positive, and False Negative samples, respectively.

% To measure the prediction performance for all four kinds of models, we choose average Accuracy@N (Acc@N), where N = 1, 10, Mean Reciprocal Rank (MRR), and Normalized Discounted Cumulative Gain (NDCG) to evaluate different models. Aside from general performance, we also want to find out the prediction performance for immobility.

\para{Experimental Settings}
%\textcolor{red}{To make a fair comparison for different types of baseline models, we use the following settings to adapt these models to the disaster mobility prediction task.
\textcolor{black}{To implement a fair comparison between different algorithms, we train LSTM, GRU, DeepMove, Flashback, and STiSAN+ on $\mathbb{D}^{N}_{T}$, train HMM+MDP and DeepMove on one part of $\mathbb{D}^{X}_{T}$, and train CHAML and CATUS on $\mathbb{D}^{X}_{S}$. For LLM4POI and ST-MoE-BERT, we directly add an instruction with the event level and train them on $\mathbb{D}^{N}_{T}$. For X-MLM, we apply it to different base models to conduct comprehensive evaluations, in which we use Llama-3-8B\footnote{https://huggingface.co/meta-llama/Meta-Llama-3-8B} as the LLM of our framework.
Finally, for all models, we conduct testing in the other parts of $\mathbb{D}^{X}_{T}$. 
}

\subsection{Overall Performance}

\begin{table*}[t]
\fontsize{2pt}{3pt}\selectfont
\setlength{\tabcolsep}{3pt}  % 减小列间距
\renewcommand{\arraystretch}{0.8}  % 增加行间距
\setlength{\arrayrulewidth}{0.1pt}  % 设置线条粗细为 0.3pt
\centering
\caption{Performance in extreme scenarios in target city. The best results are in bold and the suboptimal results are underlined.}
\begin{threeparttable}
\resizebox{0.99\textwidth}{!}
{
\begin{tabular}{l|cccccccc}
\hline
\textbf{Model} & \textbf{Acc@1} & \textbf{Acc@10} & \textbf{MRR} & \textbf{NDCG@5} & \textbf{NDCG@10} & \textbf{Pre@Immob} & \textbf{Rec@Immob} & \textbf{F1@Immob} \\ \hline
LSTM & 0.0381 & 0.1892 & 0.0436 & 0.0729 & 0.1109 & 0.1698 & 0.1375 & 0.1518 \\
GRU & 0.0367 & 0.1649 & 0.0359 & 0.0682 & 0.0903 & 0.1624 & 0.1334 & 0.1465\\
DeepMove@LSTM & 0.1354 & 0.2796 & 0.1832 & 0.1917 & 0.3198 & 0.3358 & 0.3297 & 0.3325 \\
Flashback@LSTM & 0.0854 & 0.2143 & 0.1167 & 0.1633 & 0.2874 & 0.2781 & 0.2219 & 0.2468 \\
STiSAN+ & 0.1872 & 0.3198 & 0.2187 & 0.2465 & 0.3650 & 0.3681 & \underline{0.3819} & \underline{0.3749} \\ 
HMM+MDP & 0.1532 & 0.3127 & 0.1970 & 0.2166 & 0.3319 & 0.3613 & 0.3597 & 0.3605 \\
DeepMob & \underline{0.2180} & \underline{0.4290} & \underline{0.2501} & 0.2789 & \underline{0.3701} & \underline{0.3871} & 0.3540 & 0.3699 \\
CHAML & 0.1377 & 0.2097 & 0.1224 & 0.1401 & 0.2998 & 0.2412 & 0.2588 & 0.2496 \\
CATUS & 0.1781 & 0.2297 & 0.1402 & 0.1621 & 0.3176 & 0.2776 & 0.2809 & 0.2792 \\
LLM4POI & 0.1186 & 0.2413 & 0.1522 & 0.1612 & 0.2976 & 0.2971 & 0.3087 & 0.3031 \\
ST-MoE-BERT & 0.2119 & 0.4187 & 0.2478 & \underline{0.3198} & 0.3699 & 0.3387 & 0.3290 & 0.3337 \\ \hline
X-MLM@Flashback & 0.1579 & 0.3342 & 0.1398 & 0.2022 & 0.3376 & 0.4207 & 0.4398 & 0.4301 \\
X-MLM@DeepMove & 0.2587 & 0.4982 & 0.2905 & 0.3097 & 0.3578 & 0.4863 & 0.4901 & 0.4882 \\
X-MLM@STiSAN+ & \textbf{0.2897} & \textbf{0.5118} & \textbf{0.3209} & \textbf{0.3283} & \textbf{0.4601} & \textbf{0.5012} & \textbf{0.5114} & \textbf{0.5062} \\ \hline
\end{tabular}
}
\end{threeparttable}
\label{tab:overall_performance}
\end{table*}

The performance comparison of all models is shown in Table~\ref{tab:overall_performance}.
% Overall, X-MLM with STiSAN+ achieves the best performance across all metrics.
In comparison with the best-performing baseline methods,
X-MLM with STiSAN+ as the base model shows remarkable enhancements,
with a 32.8\% improvement in Acc@1, a 19.3\% improvement in Acc@10,
a 28.3\% improvement in MRR,
a 2.6\% improvement in terms of NDCG@5,
a 24.3\% improvement in NDCG@10,
a 29.4\% improvement in Pre@Immob,
a 33.9\% improvement in Rec@Immob,
and a 35.0\% improvement in F1-score@Immob.
The key reason is that our method takes the impact of extreme events on human mobility patterns into account and we explicitly model the immobility in extreme scenarios.
We stimulate the power of LLM in understanding how extreme event affects human mobility patterns by using a hybrid tokenize strategy.
Moreover, X-MLM transfers cross-city knowledge at the intention level instead of directly transferring the entire mobility trajectory, thus achieving more accurate prediction results.
\begin{table}[]
\centering
\caption{The predictive performance of our model after removing different modules. The best performance is given in bold.}
\fontsize{7pt}{8pt}\selectfont
\setlength{\tabcolsep}{2.0pt}  % 减小列间距
\begin{tabular}{l|ccccc}
\hline
\textbf{Model}                                  & \textbf{Acc@10} & \textbf{MRR} & \textbf{Pre@Immob} & \textbf{Rec@Immob} & \textbf{F1@Immob} \\ \hline
\textbf{w/o} RAG                 & 0.3148 & 0.2691 & 0.3312 & 0.3187 & 0.3248 \\
\textbf{w/o} Soft-prompt         & 0.3387 & 0.2770 & 0.3568 & 0.3248 & 0.3400 \\
\textbf{w/o} Immobility & 0.2896 & 0.2226 & 0.2197 & 0.2613 & 0.2387 \\
\textbf{w/o} LLM Refining  & 0.3166 & 0.2531 & 0.2769 & 0.2903 & 0.2576 \\
X-MLM                          & \textbf{0.5118} & \textbf{0.3209} & \textbf{0.5012} & \textbf{0.5114} & \textbf{0.5062} \\\hline
\end{tabular}
\label{tab:ablation_study}
\end{table}
\subsection{Impact of Base Models in Different Sceanrios}
In this part, we evaluate the capability of X-MLM to capture shifts in human mobility during extreme scenarios and compensate for performance degradation across different base models.
First, we directly use base models to make predictions on trajectories in normal scenarios.
Second, we apply base models for human mobility prediction in extreme scenarios.
Third, we integrate base models with intention modulation to make predictions in extreme scenarios.
The results shown in Table~\ref{tab:intention_modulation} illustrate that the performance of different baseline methods decreases significantly in extreme scenarios,
for up to 82.2\% in terms of Acc@1, 65.2\% in Acc@10, and 61.4\% in MRR.
% This is mainly caused by the shift of mobility patterns from normal to extreme scenarios,
% which is challenging for conventional models to handle.
% In contrast, the proposed X-MLM framework compensates for such performance drops by 
% fine-tuning the LLM on the intention level to acquire knowledge of how different disasters affect human mobility patterns.
Once combined with the proposed intention modulation,
the performance of the conventional models in extreme scenarios will improve by up to 13.5\% in Acc@1, 12.4\% in Acc@10, and 7.8\% in MRR.

\begin{figure}[t]
  \centering
  % 第一行，两个子图并排
  \begin{subfigure}[b]{0.23\textwidth}  % 调整子图宽度以适应一行内两个子图
    \includegraphics[width=\textwidth]{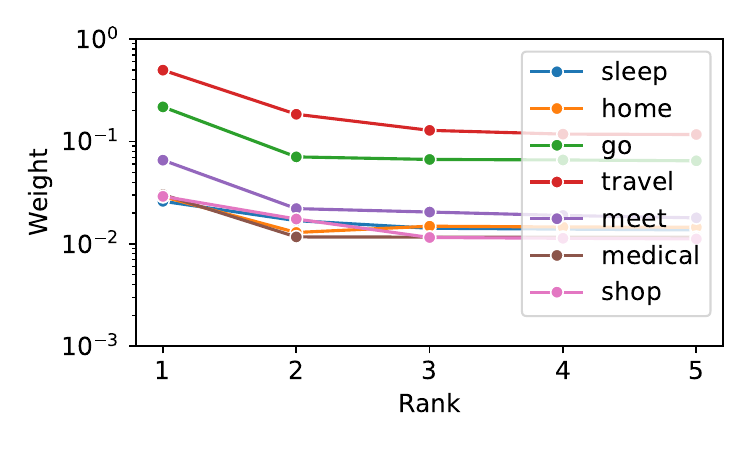}
    \caption{\small Weights for intention vocabulary examples.}
    \label{fig:intention_clip:related}
  \end{subfigure}
  \hfill  % 添加水平填充以保持间距
  \begin{subfigure}[b]{0.23\textwidth}
    \includegraphics[width=\textwidth]{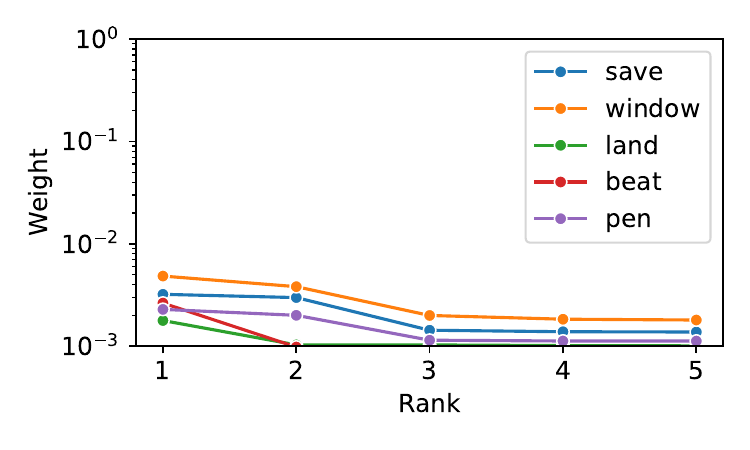}
    \caption{\small Weights for non-intention vocabulary examples.}
    \label{fig:intention_clip:non_related}
  \end{subfigure}
  \caption{Top 5 weights of vocabularies in prototypes.}
  \label{fig:intention_clip}
    % 调整垂直间距
\end{figure}
\subsection{Intention Analysis}
\iffalse
In X-MLM, we use the intention translator and predictor $\mathcal{E}_\mathbf{I}$ to predict the next intention and use LLM to refine it, especially the immobility. To measure the performance at the intention level, we conduct the following experimental studies.
\fi
\begin{figure}[t]
  \centering
  % 第一行
  \begin{subfigure}[b]{0.235\textwidth}  % [b] 表示基线对齐, {0.45\textwidth} 是子图宽度
    \includegraphics[width=\textwidth]{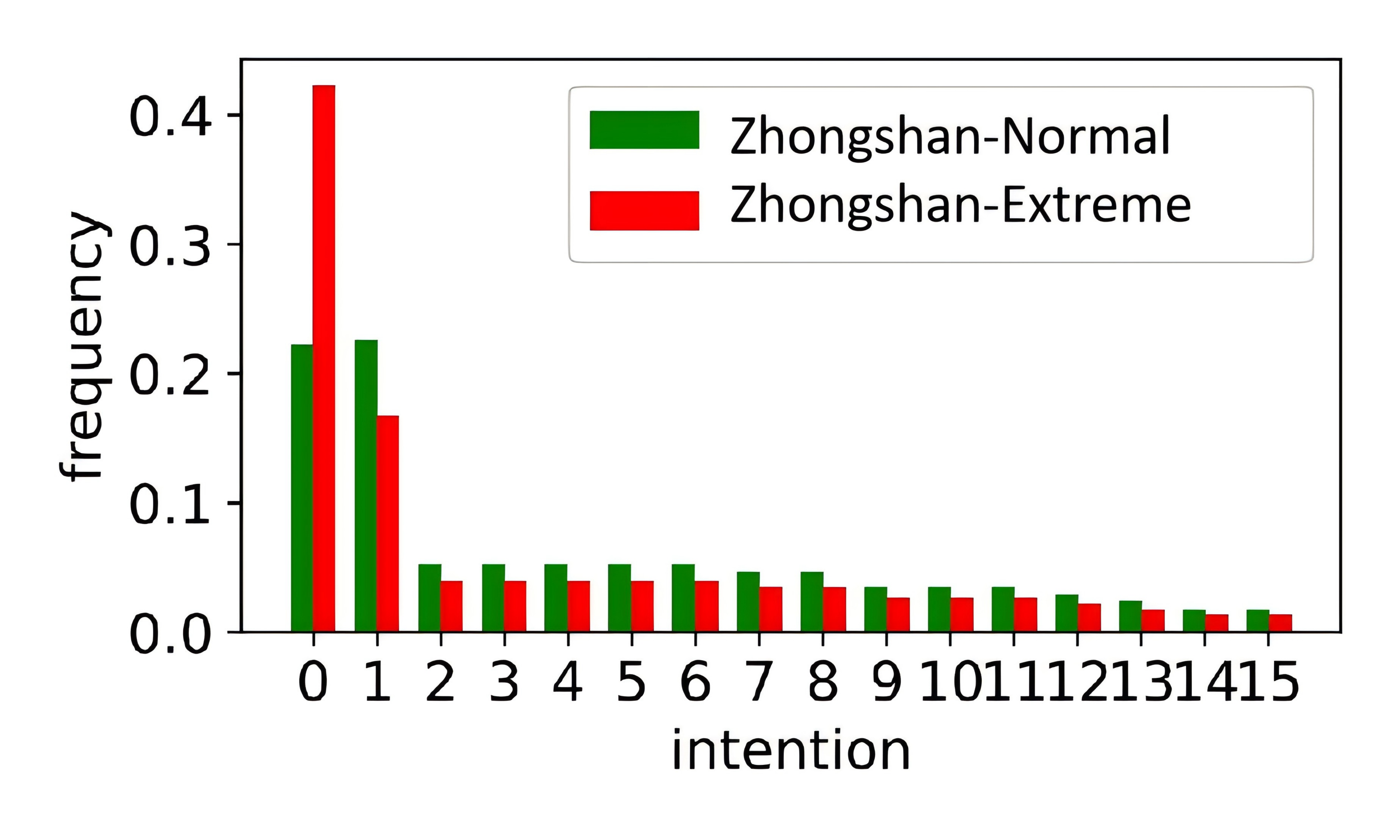}
    \caption{Intention distribution in Zhongshan.}
    \label{fig:intention_dis:a}
  \end{subfigure}
  \hfill
  %\hfill  % 在两个子图之间添加水平填充以保持间距
  \begin{subfigure}[b]{0.235\textwidth}
    \includegraphics[width=\textwidth]{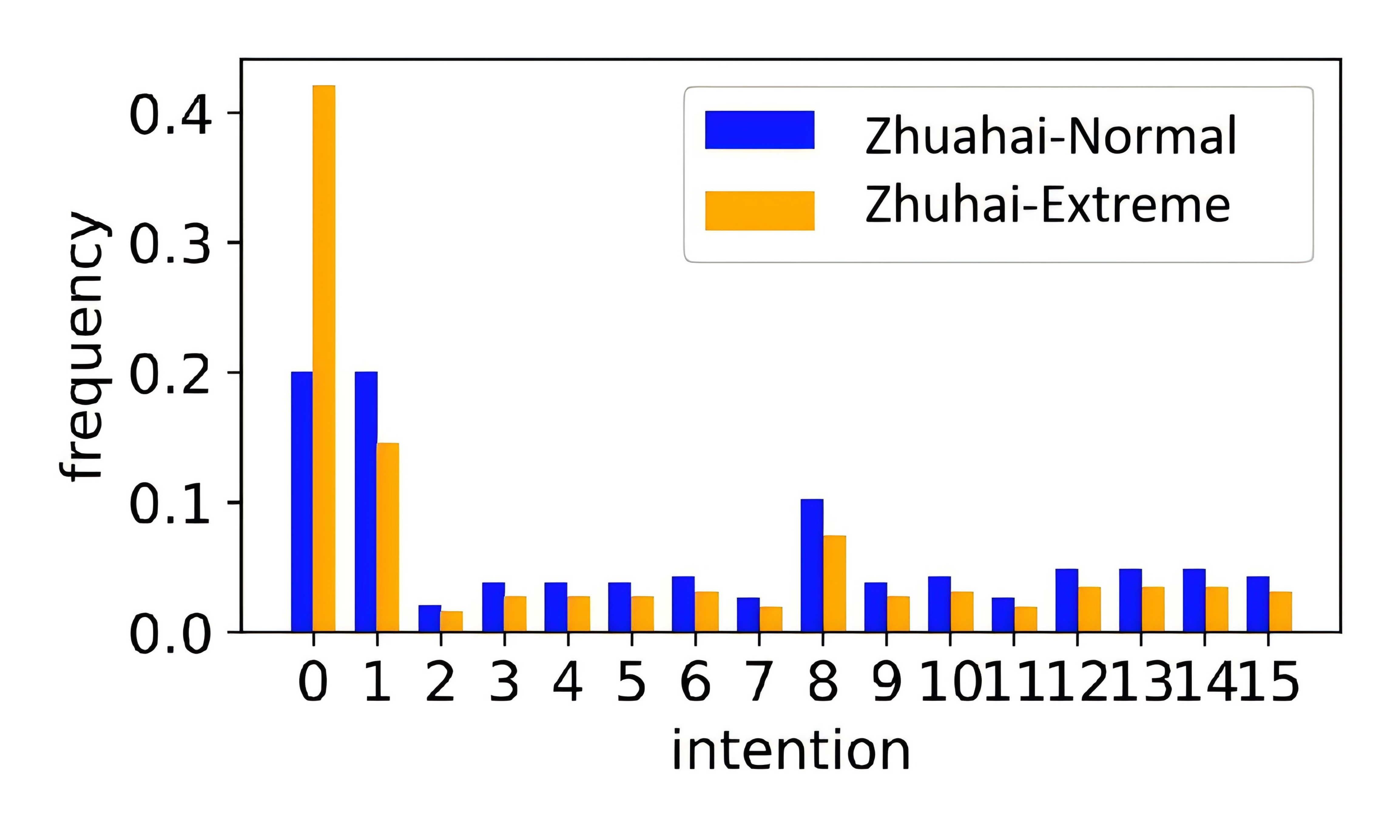}
    \caption{Intention distribution in Zhuhai.}
    \label{fig:intention_dis:b}
  \end{subfigure}
  \hfill
  \begin{subfigure}[b]{0.235\textwidth}
    \includegraphics[width=\textwidth]{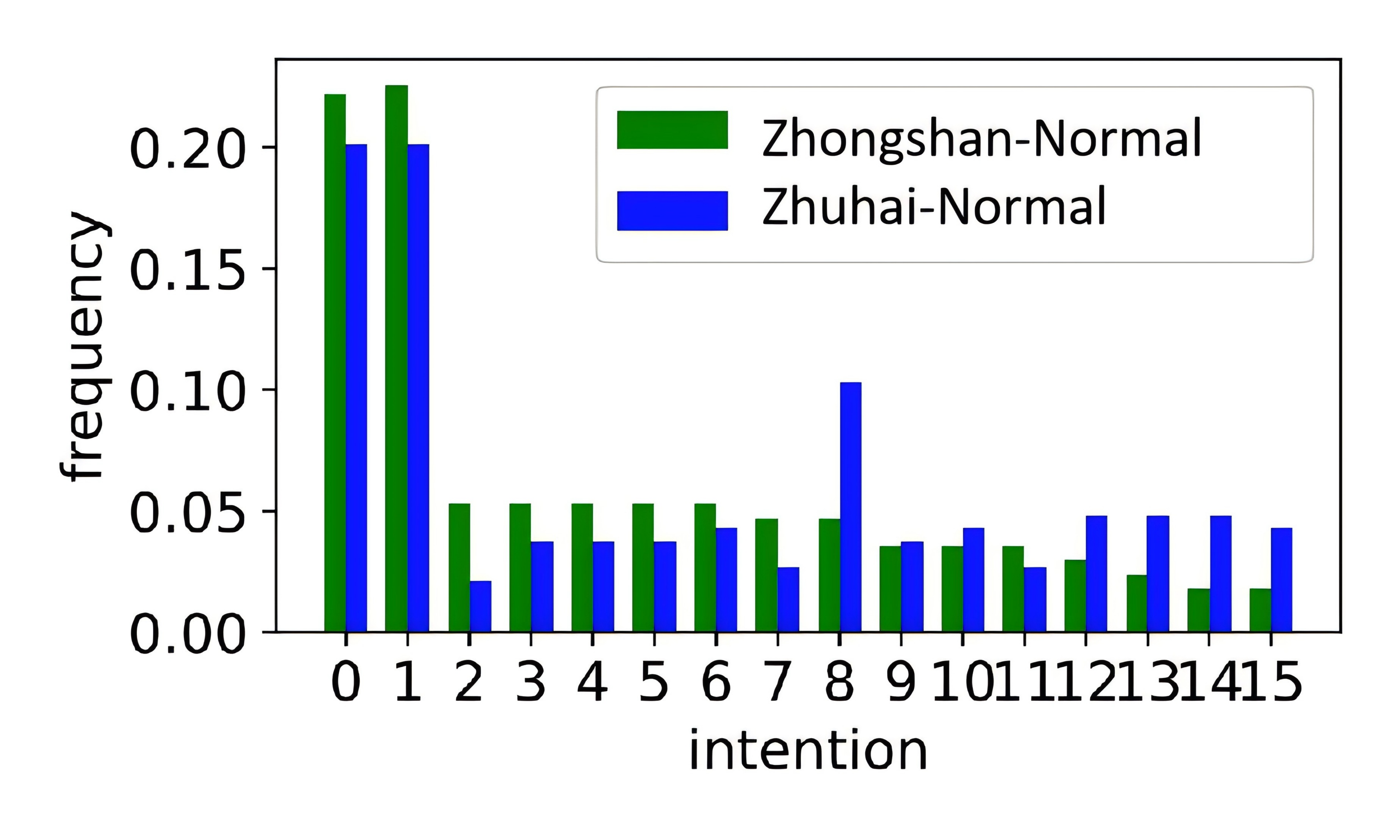}
    \caption{Intention distribution in normal scenarios of two cities.}
    \label{fig:intention_dis:c}
  \end{subfigure}
  \hfill
  %\hfill  % 在两个子图之间添加水平填充以保持间距
  \begin{subfigure}[b]{0.235\textwidth}
    \includegraphics[width=\textwidth]{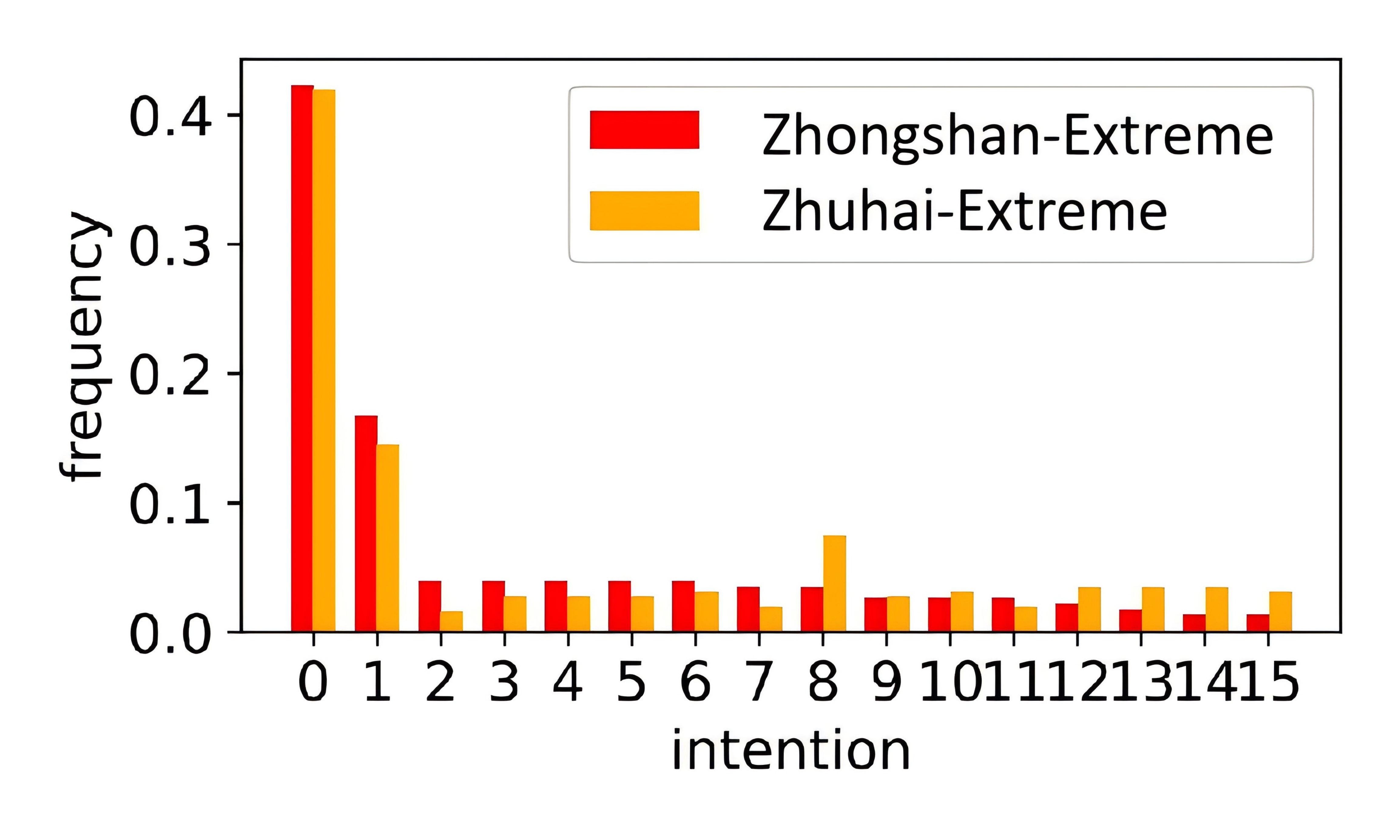}
    \caption{Comparison of intention distribution in different cities.}
    \label{fig:intention_dis:d}
  \end{subfigure}
  \caption{Comparison of intention distribution between cities.}
  \label{fig:intention_compare}
\end{figure}

\para{Intention-CLIP Analysis}
To evaluate that our Intention-CLIP can train the intention translator and predictor to learn the intention knowledge in the alignment process, we select both intention-related words (i.e., "go" and "travel") and intention-unrelated words to compare top 5 weights for different prototypes~\cite{yuan2014human}.
Intuitively, the linear layer between vocabularies and prototypes can be regarded as a matrix, which indicates how much a vocabulary contributes to different prototypes.
Thus, we select the top-5 contributions of intention-related and intention-not-related vocabularies.
The results in Fig.~\ref{fig:intention_clip} show that the top-5 intention-related vocabularies have higher weights than those intention-unrelated vocabularies.
Moreover, the top-1 vocabulary's weight is significantly higher than the weight of the second-top vocabulary, showing that the Intention-CLIP can also distinguish different intentions reasonably.
\begin{table}[H]
\centering
\caption{Comparison of intention prediction in normal (\textbf{Norm.}) and extreme event (\textbf{Dis.}) scenarios.}
\fontsize{7pt}{8pt}\selectfont
\setlength{\tabcolsep}{2.1pt}  % 减小列间距
\begin{tabular}{l|cccc}
\hline
\textbf{Scenarios}                 & \textbf{Acc}  & \textbf{Pre@Immob} & \textbf{Rec@Immob} & \textbf{F1@Immob} \\ \hline
\textbf{Norm.}           & 0.693 &      0.472      &      0.491      &        0.481       \\
\textbf{Dis.} w/o LLM & 0.438 &      0.418      &      0.428      &        0.422        \\
\textbf{Dis.}         & 0.763 &     0.719       &      0.774      &       0.745         \\ \hline
\end{tabular}
\label{tab:intention_acc}
\end{table}
\para{Intention Prediction Accuracy}
% X-MLM predicts the next location by converting it into an intention-level task with the intention translator and predictor, and the intention will be refined by using an LLM.
To evaluate the performance of the intention translator and predictor and the performance of LLM refinement \emph{at intention level}, we compare the performance of intention prediction in both normal and extreme scenarios. The results are shown in Table~\ref{tab:intention_acc}, where the intention prediction accuracy will decrease in extreme scenarios by 58.2\% in terms of Acc.
In addition, the prediction of immobility intention also decreases when the scenario shifts from normal to extreme event.
Nevertheless, with the help of powerful LLMs, our method shows better performance in intention prediction even in extreme scenarios while achieving desirable performance in immobility prediction,
with enhancements in Acc@1 by up to 10.1\%.

% \begin{figure}
%     \centering
%     \includegraphics[width=0.99\linewidth]{figs/intention_dis.pdf}
%     \caption{Intention distribution between different cities and different scenarios. (a) intention distribution in Zhongshan between two scenarios; (b) intention distribution in Zhuhai between two scenarios; (c) intention distribution in normal scenarios between two cities; (d) intention distribution in extreme scenarios between two cities.}
%     \label{fig:intention_compare}
% \end{figure}

\para{Intention Distribution in Different Cities and Scenarios}
To further explore how our model can successfully transfer mobility knowledge from different cities and extreme events by modeling at the intention level, we compare the intention distribution between different cities and extreme events. The results shown in Fig.~\ref{fig:intention_compare} indicate that the intention will heavily shift when an extreme event occurs. Meanwhile, the intention distribution between different cities is relatively smaller in both normal and extreme scenarios.
% For this reason, our model can transfer intention between different cities and disasters.

\subsection{Ablation Study}
We also conduct an ablation study to quantify how each module of X-MLM affects its overall performance. In the \textit{\textbf{w/o} RAG} variant, we remove the retrieval part for reference trajectories, indicating that the model will predict without external knowledge.
In the \textit{\textbf{w/o} Soft-Prompt} variant, we remove the soft prompt, indicating that the LLM can only acquire extreme event information from the prompt in a few-shot learning fashion.
We remove the special design of immobility in the \textit{\textbf{w/o} Immobility} variant.
In the \emph{\textbf{w/o} LLM Refining} variant, we remove the LLM-based Intention Refiner and directly use the intention embedding predicted by the Intention Translator and Predictor fine-tuned on the extreme event dataset as the modulation of the base model. The experimental results in Table~\ref{tab:ablation_study} indicate that deleting any module results in a performance decrease, highlighting the necessity of each module to accurately predict mobility in extreme scenarios.
% Specifically, removing the RAG part weakens the LLM's access to cross-city knowledge, decreasing prediction accuracy.
% The absence of the disaster-level-aware soft prompt also weakens performance, as does removing immobility modeling, which makes it harder for the LLM to capture shifts in mobility patterns during disasters.
% Finally, removing the LLM Refining reduces performance since the model no longer captures how disasters affect mobility.

\iffalse
Table~\ref{tab:ablation_study} shows the human mobility prediction results achieved by the above variants and X-MLM.
It is obvious that the removal of the model component will lead to performance loss at varying degrees.
Notably, removing the retrieval part will weaken the external cross-city knowledge of LLM, leading to a decrease in prediction performance.
The performance will also decrease without the event-level-awaring soft prompt.
% since LLM can only understand how disaster level affects human mobility in a few-shot paradigm from reference trajectories.
By removing the immobility modeling, the LLM can only answer the next intention directly without the immobility reference, making it harder for LLMs to capture the shift of human mobility patterns in extreme scenarios.
Furthermore, removing the LLM Refining part will also cause a performance decrease since the model can on longer understand the impact of extreme events on human mobility patterns.
\fi

\subsection{Inference Time Analysis}
\begin{table}[H]
\centering
\small
\setlength{\tabcolsep}{2.1pt}  % 减小列间距
\begin{tabular}{l|ccc}
\hline
\textbf{Model}                 & \textbf{LLM4POI}  & \textbf{ST-MoE-BERT} & \textbf{X-MLM} \\ \hline
\textbf{Time (s)}           & 5.61 &      1.03      &      5.67          \\ \hline
\end{tabular}
\caption{Comparison of the inference time between different LLM-Based Mobility Prediction Models.}
\label{tab:inference_time}
\end{table}

In order to evaluate the practicality of our algorithm, we analyze the inference time of both our model and the LLM-Based Mobility Prediction Model baselines, with hardware comprising two A100 GPUs. For our model and LLM4POI, we use the Llama-3.1-8B as the backbone while using BERT with 110 million parameters as the backbone of ST-MoE-BERT. The results in Table~\ref{tab:inference_time} indicate that although the inference time of our model is a little bit slower than the baselines, the inference time is still at the same magnitude. This indicates the practicality of our framework.

\section{Conclusion}
In this paper, we propose a mobility prediction model in a extreme scenario using cross-city knowledge and specially designed at the intention level. Extensive experiments illustrate that X-MLM can compensate for the performance decrease in extreme scenarios of normal scenario mobility prediction models. Future directions can consider testing our model on more kinds of extreme events, including epidemics, wildfires, etc. It is also possible to consider how government policies affect the mobility patterns in extreme scenarios, such as movement restriction and evacuation intervention.

\bibliographystyle{IEEEtran}
\bibliography{bare_jrnl_new_sample4}

\vfill

\end{document}